\pdfoutput=1

\documentclass[11pt]{article}

\usepackage[final]{acl}

\usepackage{times}
\usepackage{latexsym}
\usepackage{listings}
\usepackage{multirow}

\usepackage[T1]{fontenc}
\usepackage[utf8]{inputenc}

\usepackage{microtype}

\usepackage{inconsolata}

\usepackage{graphicx}

\usepackage{booktabs}
\usepackage{tabularx}
\usepackage{enumitem}
\setlist[itemize]{leftmargin=10pt, itemsep=0pt, topsep=0pt}
\setlist[enumerate]{leftmargin=15pt, itemsep=0pt, topsep=0pt}

\title{PVP: An Image Dataset for Personalized Visual Persuasion with\\Persuasion Strategies, Viewer Characteristics, and Persuasiveness Ratings}

\author{
 \textbf{Junseo Kim\textsuperscript{2}}\thanks{This research was conducted while the first author was an intern at the Graduate School of Data Science at SNU.}
 \textbf{Jongwook Han\textsuperscript{1}}
 \textbf{Dongmin Choi\textsuperscript{1}}
 \textbf{Jongwook Yoon\textsuperscript{1}}
 \textbf{Eun-Ju Lee\textsuperscript{2,3}}
 \textbf{Yohan Jo\textsuperscript{1}\thanks{Corresponding author.}}
\\
 \textsuperscript{1}Graduate School of Data Science, Seoul National University
 \\
 \textsuperscript{2}Interdisciplinary Program in Artificial Intelligence, Seoul National University
\\
 \textsuperscript{3}Department of Communication, Seoul National University
 \\
  \texttt{\{junseo.kim, johnhan00, chrisandjj, jonguki, eunju0204, yohan.jo\}@snu.ac.kr} \\
}

\begin{document}
\maketitle 
\begin{abstract}
Visual persuasion, which uses visual elements to influence cognition and behaviors, is crucial in fields such as advertising and political communication.
With recent advancements in artificial intelligence, there is growing potential to develop persuasive systems that automatically generate persuasive images tailored to individuals.
However, a significant bottleneck in this area is the lack of comprehensive datasets that connect the persuasiveness of images with the personal information about those who evaluated the images. 
To address this gap and facilitate technological advancements in personalized visual persuasion, we release the Personalized Visual Persuasion (PVP) dataset, comprising 28,454 persuasive images across 596 messages and 9 persuasion strategies. 
Importantly, the PVP dataset provides persuasiveness scores of images evaluated by 2,521 human annotators, along with their demographic and psychological characteristics (personality traits and values). 
We demonstrate the utility of our dataset by developing a persuasive image generator and an automated evaluator, and establish benchmark baselines.
Our experiments reveal that incorporating psychological characteristics enhances the generation and evaluation of persuasive images, providing valuable insights for personalized visual persuasion.\footnote{Our dataset and source code are available under the MIT license at \url{https://github.com/holi-lab/PVP_Personalized_Visual_Persuasion}.}

\end{abstract}

\section{Introduction}

\begin{figure*}
    \centering
    \includegraphics[width=0.9\linewidth]{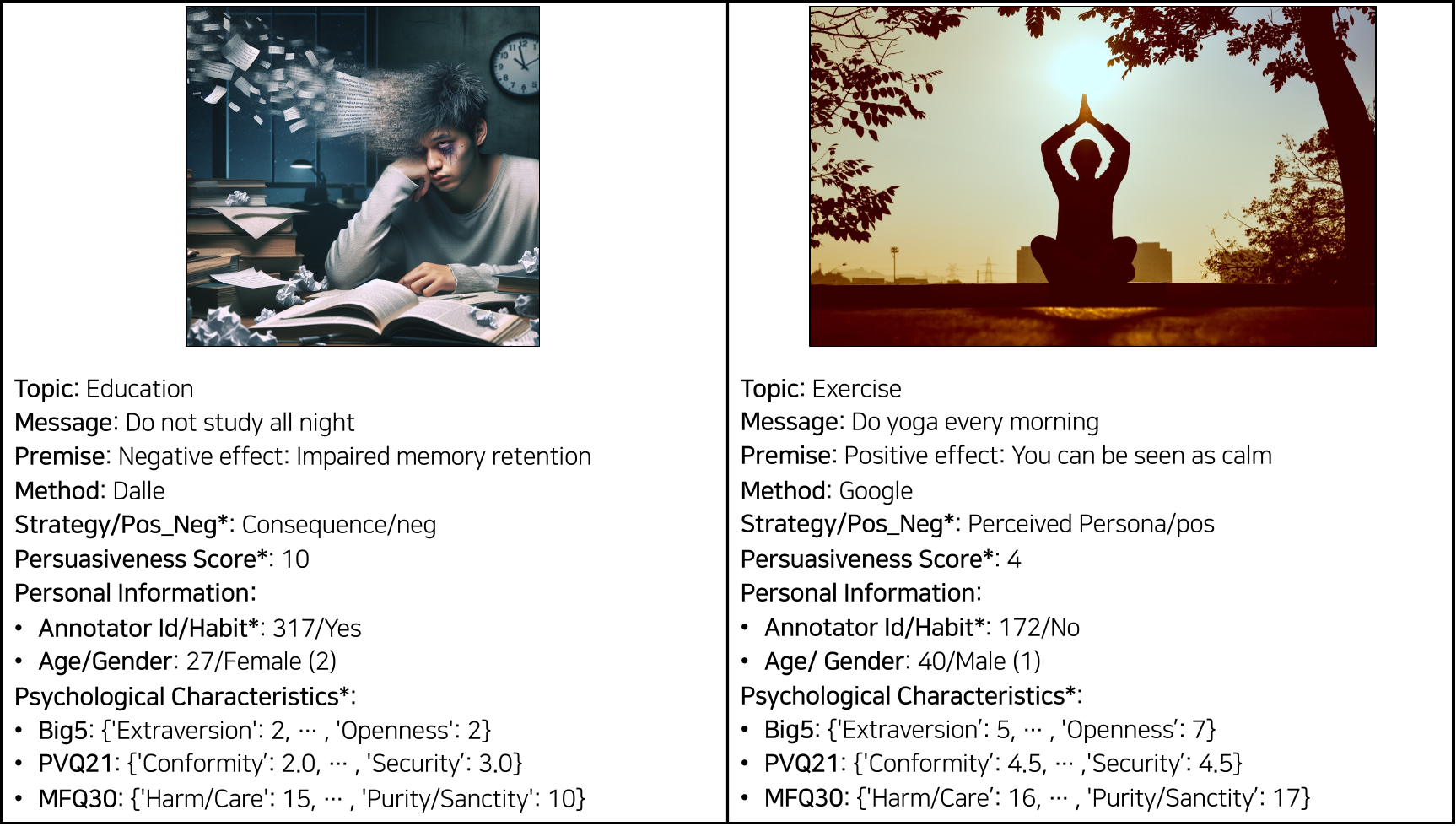}
    \caption{Examples in the PVP dataset. Asterisks (*) indicate important elements.}
    \label{fig:dataset_example}
\end{figure*}

Visual persuasion refers to using visual elements to influence cognition, emotions, and behaviors, and it plays a crucial role in fields such as advertising, memes, propaganda, and political communication~\cite{chandler2011dictionary}. Visual persuasion has been an integral part of human history, communicating power and moral values through political and religious art. Consequently, various research communities, including communication studies and social psychology, have extensively studied visual persuasion \cite{messaris1996visual,garber2003color,seo2020meta,miller1998visual}. And more recently, efforts have been made to use generative models for visual persuasion \cite{AI_visual_persuasive,kumar2023persuasion}. 

In these studies, datasets are vital for analyzing the impact and effectiveness of visual elements in communication and for training and evaluating machine learning models for persuasive systems. However, a key challenge in persuasion is that there is no ``one-size-fits-all'' approach. Despite the significant role played by the persuadee's psychological characteristics, such as personality and values, existing datasets do not provide sufficient information about such characteristics associated with the persuasion effectiveness of a given image.

To address this gap, we construct and release the \textbf{Personalized Visual Persuasion (PVP)} dataset. This large-scale dataset contains 28,454 images related to 596 messages designed to influence viewer behaviors (e.g., ``Do not smoke''), across a broad range of 20 topics based on U.S. government departments and agencies. Figure~\ref{fig:dataset_example} shows two examples in the dataset. A distinctive feature of our dataset is the use of nine \textbf{persuasion strategies} based on theoretical frameworks (e.g., gain frame: depicting a positive consequence of the target behavior). To enable images to reflect these strategies, we employed a novel method of generating images using DALLE and also sourced images through Google Image Search. 
Furthermore, recognizing that viewers' characteristics are crucial for persuasion effectiveness, we collected persuasiveness scores for the images along with the annotators' \textbf{demographics}, \textbf{habits}, \textbf{Big 5 personality traits} \cite{Big5}, and \textbf{values} \cite{schwartz2012overview,Moral_foundation_theory}.
This allows us to analyze the effectiveness of various persuasion strategies embedded in images, especially contextualized in the viewer's psychological characteristics, and to integrate this information into the development of persuasive systems. 

Empowered by the PVP dataset, we propose two tasks: (1) developing a \textbf{generative model} for creating personalized persuasive images, and (2) developing an \textbf{automated evaluator} for assessing the persuasiveness of an image tailored to the target viewer. 
Due to the importance of a consistent automated metric in developing a generator, our first experiment involved building an evaluator that predicts a given image's persuasiveness score based on a message promoting a certain behavior and the target viewer's psychological characteristics. We compared various base models and forms of input images (image itself vs. its description).
We found that the psychological characteristics of target viewers are informative in this task. 
Additionally, the effectiveness of using images as input (multimodal) versus image descriptions (text only) depends on the model; GPT-4o excelled with the former, whereas GPT-4o-mini with the latter. However, a small model fine-tuned on our PVP dataset outperformed the GPT models overall.

Our second experiment involved developing a persuasive image generator based on a message and the target viewer's psychological characteristics. We evaluated various models using our evaluator and established baseline performances as references for future work. 
Overall, our experiments revealed significant room for improvement in personalized visual persuasion, and we expect our PVP dataset to serve as a valuable resource. 

Our contributions are summarized as follows:
\begin{itemize}
\item Our work extends prior research on persuasion and argumentation, which has primarily focused on text, into the visual modality, offering opportunities to enhance the effectiveness of persuasive communication.
\item We release the first large-scale dataset, PVP, for personalized visual persuasion, containing persuasiveness scores of images with the annotators' psychological characteristics.
\item We introduce two novel tasks: generating personalized persuasive images and evaluating the persuasiveness of generated images tailored to target viewers.
\item We explore various models for these tasks, establish baseline performances, and detail our findings and suggestions for future research.
\end{itemize}

\section{Related Work}\label{sec:related_work}
In this section, we survey existing research and datasets on visual persuasion, highlighting key limitations that our study aims to overcome and the theoretical frameworks underpinning our dataset.

\begin{table*}[t]
    \small
    \centering
    \begin{tabularx}{\linewidth}{p{1.6cm}p{1.8cm}p{3.1cm}p{3.9cm}p{3.7cm}}
        \toprule
        Dataset & Image Types & Topics & Strategies & Information \\
        \midrule
        \cite{dimitrov2021semeval} 
            & Memes 
            & COVID-19, politics, vaccines, gender equality 
            & Loaded language, name calling, smears, doubt, slogans, etc.
            & Meme text, persuasion techniques \\ 
        
        \cite{liu2022imagearg} 
            & Statistics, testimony, etc. 
            & Abortion, immigration, gun control 
            & Logos, pathos, ethos
            & Persuasiveness scores, image types, persuasion strategies, tweets \\
        \cite{hussain2017automatic} 
            & ADs
            & Products, smoking, animal abuse, etc.
            & Symbolism, emotional appeal, humor, cultural references, etc.
            & Sentiment, topics, intents, persuasion strategies \\
        
        \cite{park2014computational}
            & Videos of user reviews
            & Movie reviews, general opinions
            & Verbal and non-verbal cues
            & Persuasiveness change, multimodal features \\ 
        \cite{joo2014visual} 
            & Photos of politicians 
            & Politics 
            & Emotions, trustworthy, socially dominant, favorable, gestures, etc.
            & Persuasiveness rankings, persuasion strategies, image features \\ 
        PVP (Our Dataset) 
            & Situational images 
            & 20 topics based on the U.S. executive dept. 
            & Perceived persona, internal/external emotion, consequence, bandwagon 
            & Persuasiveness scores, viewers' psychological characteristics \\ 
        \bottomrule
    \end{tabularx}
    \caption{Summary of existing datasets and their characteristics}
    \label{tab:dataset_survey}
\end{table*}

\subsection{Datasets for Visual Persuasion}

Datasets play a key role in studies of visual persuasion. Table~\ref{tab:dataset_survey} summarizes representative datasets, along with the image types, topics, persuasion strategies, and accompanying information \cite{dimitrov2021semeval,liu2022imagearg,hussain2017automatic,park2014computational,joo2014visual}. 

Despite the distinctive features of these datasets, we identify four main limitations. 
First, many datasets lack persuasiveness scores, which are essential for building persuasive systems.
Second, many datasets focus on memes and symbolism, requiring a deep level of interpretation, cultural knowledge, and accompanying text to understand the intents of the images. While rhetorically rich, these images are not ideal for everyday applications that require immediate visual impact on viewers (e.g., advertisements). 
Third, most datasets are limited to a narrow range of topics, reducing their utility to specific domains and applications, such as politics and contentious issues. 
Fourth, they fail to consider how the impact of the target viewer's psychological characteristics on persuasion outcomes. The lack of information about viewers' psychological characteristics renders them insufficient for personalized visual persuasion. 
To address these limitations, our PVP dataset covers 596 messages across 20 everyday topics and includes intuitive images with easily understood meanings. Moreover, our dataset provides rich meta-information, such as the persuasiveness scores of images and the psychological characteristics of annotators.

Along similar lines, research has explored the task of assessing image quality and aesthetics \cite{Hosu.2020,Ren.2017,Kong.2016,Yang.20220d}. However, this differs from our visual persuasion task, which inherently seeks to influence the viewer's behavior through persuasion strategies that deeply engage with the viewer's values and the argument embedded in images.

\subsection{Psychological Characteristics for Persuasion}
As personalized persuasion is more effective in inducing desired behavior changes than are non-personalized approaches \cite{orji2016personalization}, collecting annotators' psychological characteristics that might affect perceived persuasiveness is useful for developing persuasive systems. Specifically, we focus on the Big Five personality traits, values, and moral foundations.

\textbf{Big 5} evaluates an individual's personality across five major dimensions: openness, conscientiousness, extraversion, agreeableness, and neuroticism \cite{Big5}. These traits have been found to influence how susceptible individuals are to persuasive messages \cite{oyibo2017investigation}. We use a 10-item version of the Big Five Inventory (BFI-10) \cite{rammstedt2007measuring}.

\textbf{Schwartz's theory of basic values} identifies ten universal values: power, achievement, hedonism, stimulation, self-direction, universalism, benevolence, tradition, conformity, and security \cite{schwartz2006theory}. This value system has been shown to influence how individuals respond to persuasive messages with different persuasion strategies \cite{wang2019persuasion}. We use the official questionnaire with 21 items (PVQ-21) \cite{schwartz2012refining}.

\textbf{Moral foundations theory} categorizes human values into five fundamental principles that influence decision-making: care, fairness, loyalty, authority, and sanctity \cite{feinberg2019moral,voelkel2018morally}. We use a 30-item questionnaire (MFQ-30) \cite{MFQ-30}.

\subsection{Persuasion Strategies}
In visual persuasion, images appeal to various cognitive and emotional aspects, employing different strategies to influence viewers. Studies in psychology and communication have extensively examined the effectiveness of various strategies in shaping the audience's opinions and behaviors. 
For instance, according to the theory of planned behavior, persuasion effectiveness depends on factors such as self-efficacy for performing the behavior, whether the target behavior is perceived as beneficial or harmful, and how it is viewed by important people in one's life \cite{ajzen1985intentions,ajzen1987attitudes}.
Modern argumentation theory has identified common argumentation schemes used in everyday arguments, such as argumentation from consequences or popular opinion \cite{walton2008argumentation}. 
Positive and negative framing emphasize the benefits of an action or the drawbacks of inaction \cite{tversky1981framing}. To that end, we carefully curate images in our dataset to represent various persuasion strategies drawing upon these theoretical frameworks.

\subsection{Generator and Evaluator for Persuasion}

Recent studies have introduced new datasets in the field of persuasive text generation to enhance the effectiveness of language. For instance, \citet{singh2024measuring} delve into the transformation of non-persuasive text into persuasive counterparts, presenting a robust framework for evaluating the efficacy of these transformations. Such research underscores the increasing focus on computational approaches to persuasion, particularly within text-dominant domains.

Despite these advancements, existing text-based datasets predominantly emphasize linguistic features of persuasion, overlooking the critical role of visual elements. This limitation is especially significant in real-world applications like advertising and public health campaigns, where visuals play a pivotal role in influencing audiences. To address this gap, our PVP dataset integrates both textual and visual modalities, providing a comprehensive resource for investigating persuasion strategies and evaluating their effectiveness across diverse contexts. This multimodal approach opens new avenues for exploring the interplay between text and visuals in persuasive communication.

\section{Personalized Visual Persuasion (PVP) Dataset}
\label{sec:dataset}
\begin{figure*}[t]
    \centering
    \fcolorbox{black}{white}{\includegraphics[ width=0.98\linewidth]{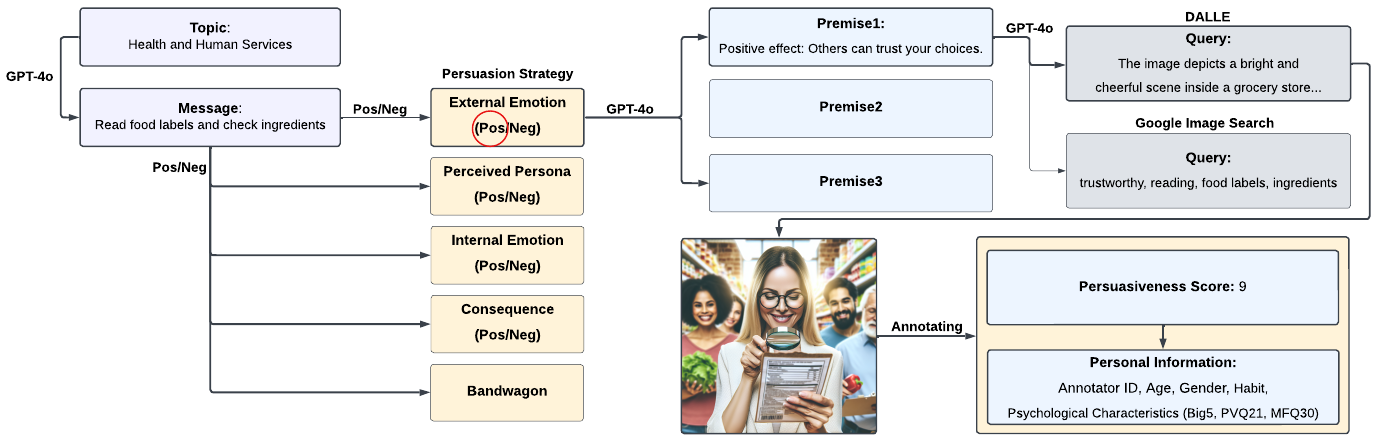}}
    \caption{Data construction process. See the main text for details.}
    \label{fig:data_construction_1}
\end{figure*}

Our PVP dataset consists of six primary elements: messages, persuasion strategies, premises, images, persuasiveness scores, and psychological characteristics of annotators. This section outlines the dataset's key components and the construction process, as illustrated in Figure~\ref{fig:data_construction_1}.

\subsection{Messages}
A message is a target behavior we want the viewer to adopt (e.g., ``Do not smoke''). To gather a diverse set of messages widely relevant to the general public, we began by identifying 15 topics based on the 15 executive departments of the United States, as these departments formulate policies across essential areas of our everyday living. 
Additionally, we identified and included 5 underrepresented topics based on U.S. government agencies. 
For each of the 20 topics, we compiled 28--30 concrete and actionable target behaviors (i.e., messages) using GPT-4o (Appendix~\ref{app:Messages}). 
Table~\ref{tab:topics} provides their descriptions and example messages.

\subsection{Persuasion Strategies}
Incorporating diverse and effective persuasion strategies into images is crucial for the utility of our dataset. Based on the theoretical frameworks outlined earlier, we have adopted the following strategies, each emphasizing a different aspect of viewer engagement with the target behavior:
\begin{itemize}
    \item \textbf{Perceived Persona}: How the viewer's persona or attributes would be perceived by others.
    \item \textbf{Internal Emotion}: Emotional reactions the viewer may personally experience.
    \item \textbf{External Emotion}: Emotional responses that other people may experience.
    \item \textbf{Consequence}: Consequences other than perceived persona and emotional responses (e.g., harms, wealth).
    \item \textbf{Bandwagon}: How popular the target behavior is among other people. 
\end{itemize}

In addition, positive and negative framing are key factors in the effectiveness of persuasion \cite{mads_nordmo__2015}. Therefore, we have further broken down the first four strategies into positive and negative frames. The positive version (i.e., gain frame) emphasizes the beneficial outcomes of engaging in the target behavior, while the negative version (i.e., loss frame) highlights the adverse outcomes of avoiding the target behavior. This approach results in a total of nine strategies.

\subsection{Premises}
A premise provides a specific reason or evidence reflecting a persuasion strategy. For instance, for the message ``Do not smoke'' and the strategy of positive external emotion, a potential premise could be ``others will appreciate a smoke-free environment''. We generated three premises for each persuasion strategy using GPT-4o, resulting in a total of 27 premises per message (Appendix~\ref{app:Pemise}).

\subsection{Queries}
To obtain images that convey specific premises using DALLE and Google, we converted each premise into a suitable prompt for DALLE and a search query for Google using GPT-4o (Appendix~\ref{app:Queries}). Figure~\ref{fig:data_construction_1} illustrates example queries.

\subsection{Images}
Using the queries above, we collected one image from DALLE and one from Google Image Search for each premise, resulting in 54 images for each message (9 strategies × 3 premises × 2 sources). Ensuring that each image clearly communicates the intended premise is essential for post-hoc analysis of the associations between persuasion strategies and persuasiveness scores, as well as for developing persuasive systems. To achieve this, we implemented a thorough filtering process involving both human and GPT evaluations. Additionally, we filtered out text-heavy images from Google, as they rely more on verbal content than visual elements and may not be useful for people who speak different languages. The validation process is detailed in Appendix~\ref{app:Images}. On average, 6 images per message were discarded, resulting in 28,454 images in the dataset.\footnote{For Google images, we will release only their URLs to avoid licensing issues.}

\subsection{Persuasiveness Scores}
Annotators were assigned a message and instructed to rate each image on a scale from 0 to 10 based on how strongly the image motivated them to adopt the behavior.\footnote{Rating perceived persuasiveness, rather than tracking actual behavior change, has been widely adopted in visual persuasion datasets \cite{liu2022imagearg, park2014computational, joo2014visual}, and aligns with established literature that uses self-reported ratings as reliable proxies for persuasive impact \cite{ajzen1991theory,webb2006does}. For our dataset, measuring behavior change across 596 messages also poses logistical and ethical barriers. We believe our rating method is effective in identifying persuasive image features and facilitating their integration into AI development.} We implemented a rigorous mechanism to detect and filter out invalid annotations (Appendix~\ref{app:Annotation Validation}). 

To minimize sampling bias and avoid the disproportionate influence of certain annotators, we partnered with a Korean survey company to recruit annotators evenly across genders and age groups (20s, 30s, 40s, and 50s). Each annotator participated in only one message (i.e., 54 images maximum). Since one of our main objectives is to address the subjectivity in assessing the persuasiveness of images, each image was rated by four different annotators.\footnote{Our annotators are Korean, which we believe contributes to greater cultural diversity in our research field, where many public datasets predominantly reflect English- or Chinese-speaking cultural contexts. Our data collection protocol is not constrained to a specific culture, however, incorporating universally established psychological traits. We leave data collection from additional cultures to future work.}

\subsection{Psychological Characteristics}
After rating persuasiveness, annotators completed three questionnaires to profile their personalities and values: the Big Five Inventory (BFI-10), the Portrait Values Questionnaire (PVQ-21), and the Moral Foundations Questionnaire (MFQ-30). In addition, since the perceived persuasiveness of an image is likely to vary depending on whether the viewer is already engaging in the behavior, annotators were asked if they were practicing the behavior on a daily basis (Habit).\footnote{This study was approved by our institution's IRB.}

Consequently, our dataset includes annotations from a total of 2,521 annotators, ensuring high diversity. Each annotator was paid \$2.90, which aligns with the minimum wage rate (\$6.87) and the task completion time (24 min). Refer to Appendix~\ref{app:Annotation Validation} for the validity of the annotations.


\section{Dataset Analysis} 
This section describes the results of the analysis of the PVP dataset, emphasizing key observations regarding image persuasiveness and the influence of topics and psychological characteristics.

\subsection{Basic Statistics}
The distribution of persuasiveness scores within the dataset forms a bell curve centered around a mean of 4.65, with noticeable peaks at the extremes (0 and 10). 
Different age groups showed slightly different modes with older groups tending to assign higher persuasiveness scores (Figure~\ref{fig:Age}). Gender differences were subtle, although male annotators tended to award slightly higher scores (Figure~\ref{fig:Gender}).

\subsection{Topics and Messages} 
\label{app:Topics and Messages}
Table~\ref{tab:average_persuasiveness_scores_topic} presents the average persuasiveness scores across different topics. Topics such as transportation, interior, and homeland security received the highest persuasiveness scores, while defense, treasury, and cyber etiquette scored the lowest. These findings suggest that images representing behaviors that are easy to adopt and deemed necessary (e.g., safety and environmental protection) are rated more favorably. For instance, the message ``Reduce your speed when there are many pedestrians'' achieved the highest rating (8.19). Conversely, messages advocating behaviors that demand significant efforts or are rather idiosyncratic (e.g., military education and personal finance management) garnered low scores. For instance, the message ``Do squats'' recorded the lowest score (1.70). These results align with the planned behavior theory \cite{ajzen1991theory}---confidence in one's ability to perform a behavior is a key factor for persuasiveness.

Annotators who had already adopted the target behavior tended to rate the images more favorably (5.0) than those who had not (4.3) (Figure~\ref{fig:habit_yes_no.png}). This phenomenon can be interpreted as an attempt to avoid cognitive dissonance, confirming the value of their habit (or a lack thereof).

\subsection{Persuasion Strategies}
\begin{figure}[t]
    \centering
    \includegraphics[width=1\linewidth]{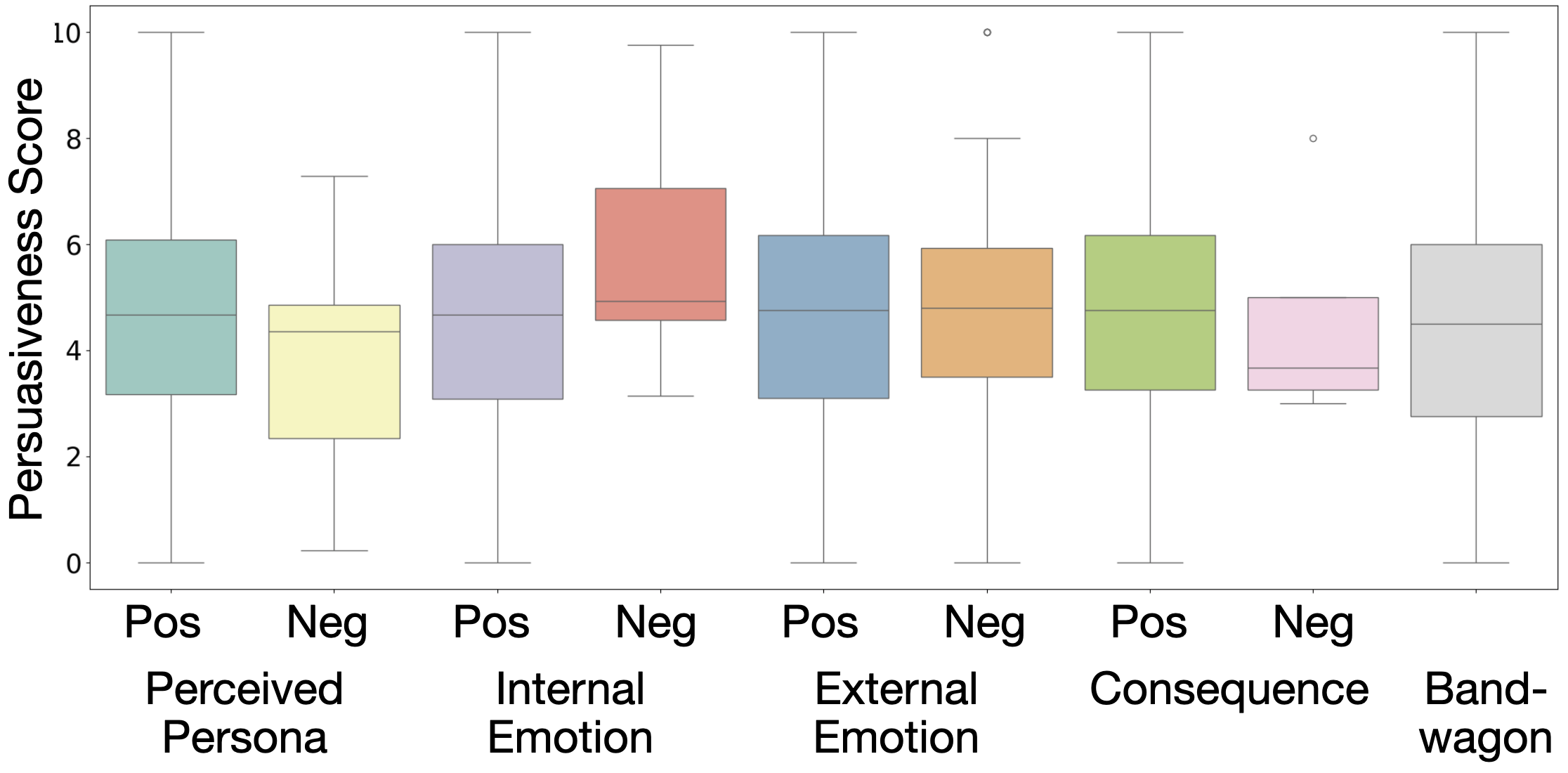}
    \caption{Persuasiveness scores by persuasion strategy.}
    \label{fig:strategy}
\end{figure}

Our analysis of persuasion strategies provides more nuanced insights beyond mere topical influences. Figure~\ref{fig:strategy} displays the distribution of persuasiveness scores for different persuasive strategies. The strategy of negative internal emotion yielded the highest average score (5.83), while the strategy of negative perceived persona garnered the lowest average score (3.73). These results suggest that direct appeals to viewers' emotions are generally more effective than those involving potential threats to public images. Interestingly, positive persuasion strategies tended to achieve higher persuasiveness scores than negative strategies overall (Figure~\ref{fig:strategy_type}). This may be attributed to the tendency of positive frames to inspire and uplift the viewer, more likely to engage and motivate them effectively.

\subsection{Personality and Values}
Incorporating viewers' psychological characteristics, such as personality traits and values, into persuasion strategies enhances our understanding of the dynamics in visual persuasion. To examine how various persuasion strategies interact with different personality traits and values, for each strategy, we calculated the correlation between personality or value scores and the persuasiveness scores of images associated with that strategy (Appendix~\ref{app:Personality and Values}).

Positive persuasion strategies exhibit consistently weak correlations across personality traits and values, suggesting that such strategies receive relatively uniform evaluations regardless of an individual’s underlying psychological characteristics.

In contrast, the negative consequence strategy exhibits significant correlations with several traits and values. It correlates positively with hedonism (0.63), suggesting that individuals who value pleasure and comfort are more responsive to messages emphasizing the negative outcomes of avoiding the target behavior. Conversely, its negative correlations with stimulation (\textminus1.00) and harm/care (\textminus0.89) suggest that the loss-framed strategy tends to have a counterproductive effect on individuals motivated by novelty seeking or moral concern.

The strategy of negative internal emotion shows a moderate positive correlation with neuroticism (0.57) and a moderate negative correlation with conscientiousness (\textminus0.51).
These patterns suggest that emotionally unstable individuals are more sensitive to internalized negative emotions such as guilt or anxiety, whereas those high in self-discipline and orderliness are less influenced by them.

More results are presented in Figure~\ref{fig:correlation_trait}. Some notable correlations between psychological characteristics and certain topics are described in Appendix~\ref{app:Correlations between Personality/Values and Topics}.

\subsection{DALLE vs. Google Image Search}
Deciding whether to generate persuasive images with an AI model or retrieve them from the web is a significant consideration. Our analysis reveals that, compared to Google images, DALLE images align better with intended premises and received slightly higher persuasiveness scores (Figure~\ref{fig: Dalle_Google}). This highlights the promise of image generation models as valuable tools for crafting personalized visual persuasion strategies. Further discussion is provided in Appendix~\ref{app:DALLE vs. Google Image}.

\section{Experiments}
\label{app:Experiments_main}


Our goal is to build a persuasive system that produces persuasive images tailored to the target viewer. To that end, we introduce two novel tasks and establish baselines leveraging our PVP dataset. First, considering that an automated metric would facilitate the development of an image generation system and provide consistent measurements across experiments, the first proposed task is to build an \textbf{evaluator model} that assesses a given image's persuasiveness score. 
The second task is to build a \textbf{generator model} that generates a personalized persuasive image.

For the development of both the evaluator and generator models, we randomly selected one annotation per image for validation, one for testing, and allocated the remaining annotations to the training set. Consequently, our training set includes 63,423 annotations, while the validation and test sets contain 28,454 annotations each.

\subsection{Evaluator}
Our evaluator model takes as input a message, an image, and the target viewer's psychological characteristics in textual format (e.g., ``Benevolence: 6.0, Conformity: 3.5, ...'') from the test set. It outputs a persuasiveness score for the image, ranging from 0 to 10. Details are provided in Appendix~\ref{app:Evaluator}.

\subsubsection{Models}
We experimented with three representative commercial and open-source models: GPT-4o, GPT-4o-mini, and LLaMA3-8B-Instruct. To analyze an effective format of input images, we compared the use of direct images (multimodal) vs. their descriptions (text-only) obtained from GPT-4o. Additionally, we explored zero-shot prompting and fine-tuning.


\subsubsection{Evaluation Metrics}
To evaluate the evaluators, we use metrics such as Spearman correlation, Pearson correlation, NDCG, and RMSE. They measure how closely an evaluator model's predicted persuasiveness scores align with the scores provided by the actual target users in the test set.

\subsubsection{Results}
\begin{table}[t]
    \small
    \centering
    \begin{tabularx}{\linewidth}{@{}p{1.2cm}cccc}
        \toprule
         & Spearman$\uparrow$ & Pearson$\uparrow$ & NDCG$\uparrow$ & RMSE$\downarrow$ \\ \midrule
        PVQ-21 & \textbf{0.25} & \textbf{0.25} & 0.42 & \textbf{3.40} \\
        Big5    & 0.23     & 0.24    & \textbf{0.43} & 3.66     \\
        MFQ-30  & \textbf{0.25}    & 0.24    & 0.42 & 3.48     \\
        None & 0.23 & 0.23 & 0.43 & 3.78 \\ \bottomrule
    \end{tabularx}
    \caption{Evaluator performance across different input psychological characteristics.}
    \label{tab:evaluator_ablation}
\end{table}

\paragraph{Psychological Characteristics}
We first examined the impact of each psychological category on the accuracy of predicting persuasiveness scores.
For this experiment, we fine-tuned four evaluators, each incorporating: \textbf{(1) values (PVQ-21)}, \textbf{(2) personality traits (Big5)}, \textbf{(3) moral foundations (MFQ-30)}, and \textbf{(4) None} (the evaluator is trained without psychological characteristics). The prompts used for fine-tuning each model are provided in Appendix~\ref{app:Prompts for Evaluator with Four Types of Input Characteristics}, and further implementation details for the fine-tuning process are detailed in Appendix~\ref{app:Implementation Details for Evaluator}.

Table~\ref{tab:evaluator_ablation} presents the results. Incorporating psychological characteristics (rows 1--3) enhances prediction accuracy compared to excluding this information (row 4), with PVQ being most effective.

\begin{figure}[t]
\centering
\includegraphics[width=1\linewidth]{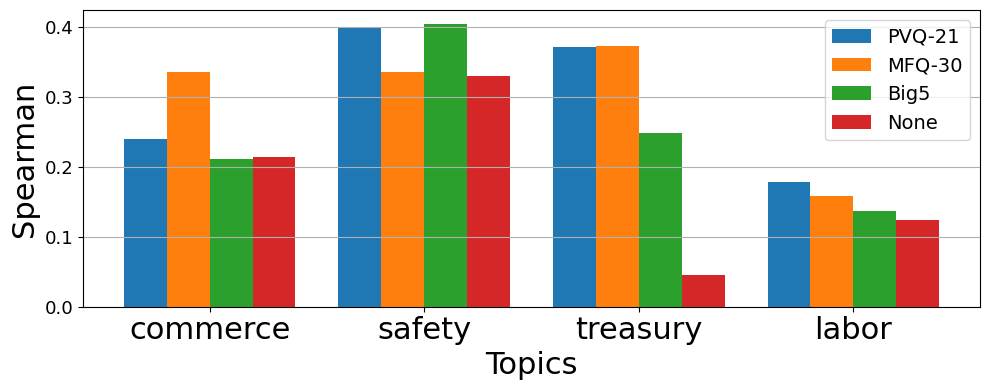} 
\caption{Spearman correlations between evaluator scores and human judgments across topics.}
\label{fig:metric_per_topic}
\end{figure}

It is reasonable to expect that the influence of psychological characteristics on persuasiveness varies across topics.
To examine this relationship, we analyzed the Spearman correlation of the four models by topic. 
A key finding is that incorporating psychological characteristics improves the model's alignment with human scores for topics that are highly relevant to those characteristics (Figures~\ref{fig:metric_per_topic} and \ref{fig:full_spearman_per_topic}).
For example, the treasury topic includes messages related to financial management, which is heavily influenced by an individual's values and personality. As a result, incorporating psychological characteristics substantially enhances prediction accuracy. 
In contrast, the safety topic features messages universally recognized as important, making prediction accuracy less sensitive to input psychological characteristics.
A more detailed analysis is provided in Appendix~\ref{app:Spearman Correlations by Topic}.

Since the PVQ-21 model demonstrated the best overall performance among the four models, we use PVQ-21 as the default setting for psychological characteristics in the subsequent experiments.

\begin{table}[t]
    \small
    \centering
    \begin{tabularx}{\linewidth}{@{}p{2cm}*{4}{>{\centering\arraybackslash}X}@{}}
        \toprule
        Model & Spearman$\uparrow$ & Pearson$\uparrow$ & NDCG$\uparrow$ & RMSE$\downarrow$ \\ \midrule
        \multicolumn{3}{l}{\hspace{-2mm}\textit{Images as input}} \\
        GPT-4o & 0.19  & 0.19  & 0.39  & 3.90  \\
        GPT-4o-mini & 0.13 & 0.11 & 0.35 & 4.01 \\
        \midrule
        \multicolumn{3}{l}{\hspace{-2mm}\textit{Image descriptions as input}} \\
        GPT-4o & 0.16 & 0.17 & 0.37 & 3.81 \\
        GPT-4o-mini & 0.15 & 0.13 & 0.36 & 3.84 \\
        LLaMA3-8B-Z & 0.07 & 0.06 & 0.34 & 3.71 \\ 
        LLaMA3-8B-F & \textbf{0.25} & \textbf{0.25} & \textbf{0.42}& \textbf{3.40} \\ \bottomrule
    \end{tabularx}
    \caption{Evaluator performance across different input image formats and base models. For LLaMA3-8B-Instruct, Z and F refer to zero-shot and fine-tuning.}
    \label{tab:evaluator_results}
\end{table}

\paragraph{Input Image Formats}
Table~\ref{tab:evaluator_results} compares prediction accuracy by input image formats (direct images vs. image descriptions). 
For the GPT models (rows 1--4), GPT-4o excelled with direct images (row 1), whereas GPT-4o-mini performed best with image descriptions (row 4), likely reflecting differences in their multimodal capabilities to interpret and reason over images.

The small open-source model LLaMA3-8B-Instruct struggled with zero-shot prompting (row 5). This model failed to understand the task and tended to predict random numbers. 
However, fine-tuning LLaMA3 with our PVP dataset substantially improved its accuracy (row 6), outperforming all other models. 
Given that this improvement was achieved using a simple fine-tuning method without extensive prompt engineering, we anticipate further performance gains through more sophisticated training techniques and utilization of our dataset. We leave this exploration to future work.

\begin{table}[t]
    \small
    \centering
    \begin{tabularx}{\linewidth}{@{}Xcccc}
        \toprule
         & Spearman$\uparrow$ & Pearson$\uparrow$ & NDCG$\uparrow$ & RMSE$\downarrow$ \\ \midrule
         Full    & \textbf{0.32}    & \textbf{0.31}    & \textbf{0.66} & \textbf{4.05}  \\
        Filtered & 0.27 & 0.29 & 0.63 & 4.70 \\
         \bottomrule
    \end{tabularx}
    \caption{Evaluator performance on the filtered test set. \textbf{Full}: LLaMA3-8B-F in Table~\ref{tab:evaluator_results}, \textbf{Filtered}: LLaMA3-8B-Instruct trained on extreme images.}
    \label{tab:evaluator_extreme}
\end{table}

\paragraph{Focus on Good and Bad Images}
It is challenging to differentiate the persuasiveness of images with scores near the center of the score distribution.
Further, in practice, it is likely more important to distinguish between good and bad images.
To that end, we tested the performance of evaluators on a filtered test set that retains only images with scores of 0--2 and 8--10.
For comparison, we evaluated the original LLaMA3-8B-F (fine-tuned on the entire training set) and a variant fine-tuned only on a subset of the training data filtered in the same way.

As shown in Table~\ref{tab:evaluator_extreme}, LLaMA3-8B-F achieved significantly higher correlations on this filtered test set (row 1) compared to the entire test set (Table~\ref{tab:evaluator_results} row 6).
On the other hand, fine-tuning a model exclusively on filtered data (row 2) resulted in degraded performance.
We speculate that training a model on a larger number of images and a broader score distribution is important, likely enhancing the model's ability to understand the relative persuasiveness of images.
Beyond this finding, this experiment also showcases the versatility of our dataset for conducting various analyses tailored to different objectives.

\paragraph{Other Analyses}
The LLaMA3-8B-F model shows little to no difference in performance between DALLE images and Google images
(Appendix~\ref{app:Comparison of Metrics by Google and DALLE}).
Further, the model exhibits good generalizability to unseen messages (Appendix~\ref{app:generalizability_of_evaluators_to_unseen_messages}).

\subsection{Generator}
\begin{table}[t]
\centering
\resizebox{\columnwidth}{!}{%
\begin{tabular}{lcccc}
\toprule
Model & Average $\uparrow$ & Standard deviation \\ \midrule
GPT-4o & 4.45 & 2.41  \\
GPT-4o-mini & 4.59  & 2.30 \\ 
LLaMA3-8B-Instruct-Finetune  & \textbf{4.77}  & 2.37 \\ \bottomrule
\end{tabular}%
}
\caption{Generator performance.}
\label{tab:generator_results}
\end{table}

The goal of the generator is to create a personalized persuasive image given a message and the target viewer’s psychological characteristics. We employed our evaluator to produce an automated metric. Though its correlation with human judgments from our dataset is moderate (0.32 for images with low/high scores), it provides a consistent and deterministic measurement across experiments. This consistency is particularly crucial, as we aim to set benchmark performances for existing models as references for future research.

We designed our generator to generate \textit{image descriptions} (not images directly) that can subsequently be used as prompts for separate image generation models. This approach allows us to isolate the image generation quality of different models from their inherent ability to capture the desired properties of persuasive images. Additionally, the most effective configuration of our evaluator (explored in the previous section) uses image descriptions as input. To that end, our generator takes a message and psychological characteristics in textual format (e.g., ``Benevolence: 4.0, Conformity: 4.5, ...''), and generates an image description (e.g., ``The film director appears to ...'').

\subsubsection{Models}
We experimented with GPT-4o, GPT-4o-mini, and LLaMA3-8B-Instruct, comparing the effectiveness of zero-shot prompting against fine-tuning. We fine-tuned the LLaMA3 model on highly persuasive images in the training set with a persuasiveness score of over 8, because fine-tuning on low-quality images can rather degrade the generator’s performance. The implementation details and prompts are detailed in Appendices~\ref{app:Implementation Details for Generator} and \ref{app:Prompts for Generator}.


\subsubsection{Evaluation}
For evaluation, generators first generate image descriptions based on the messages and psychological characteristics of the annotators in the test set. Next, we use the best evaluator, LLaMA3-8B-F, to compute their persuasiveness scores. For each model, the average score and standard deviation are reported. 
Note that this evaluation procedure does not directly use the images or the persuasiveness scores annotated in the test set.

\subsubsection{Results}

\paragraph{Image Generation} 
As shown in Table~\ref{tab:generator_results}, the fine-tuned LLaMA3 performed best, followed by GPT-4o-mini and the GPT-4o, highlighting the potential utility of our dataset for developing personalized persuasive image generation systems.
A comparison of image descriptions generated by the three models is provided in Appendix~\ref{app:comparison of image descriptions}

\paragraph{Error Analysis}
We reviewed 116 image descriptions generated by the LLaMA3 model that received scores of 0, 1, or 2. Two major error types were identified. The primary error was a misalignment between the image description and the message, where the image failed to effectively convey the intended message. The second most frequent error was a misunderstanding of psychological characteristics, where the generator struggled to adequately capture key aspects of the target psychological characteristics, particularly values. Further details are provided in Appendix~\ref{app:Error Analysis of the Generator}.



\paragraph{Implications}
Evidently, there is significant room for improvement for fine-tuning generator models. It is beyond the scope of this paper to thoroughly examine advanced fine-tuning methods. 
However, we hope that the baselines we have established as well as our dataset will serve as a valuable resource for developing advanced generator models that can effectively incorporate psychological characteristics for personalized visual persuasion.

\section{Conclusion}
We release the Personalized Visual Persuasion (PVP) dataset designed to advance personalized approaches to visual persuasion. This large-scale dataset includes persuasiveness scores for images and the psychological characteristics of the evaluators, revealing a significant impact of viewer psychology on image persuasiveness. Based on these findings, we proposed two novel tasks: generating personalized persuasive images and evaluating persuasiveness tailored to viewer characteristics. We explored various models and established baseline performances. Additionally, fine-tuning a small model on our dataset demonstrated promising performance improvements. Our study offers a springboard for future research that aims to advance the efficacy of personalized visual persuasion.

\section{Limitations}
Our study relies on self-reported ratings as proxies for persuasiveness, rather than directly measuring actual behavioral changes. 
While self-reported evaluations are widely used and considered reliable in existing literature \cite{liu2022imagearg,webb2006does,ajzen1991theory}, they may not fully capture the complex relationship between perceived persuasiveness and actual behavioral change. 
In our study, however, measuring behavioral changes across 596 topics involves ethical and practical challenges. For instance, tracking behavioral changes in response to messages like ``Do not share your account number on social media'' would require prolonged observation, raising privacy and ethical concerns.
To complement our findings, we plan to conduct studies about actual behavioral outcomes on a few topics.
These additional studies aim to validate and extend the findings of this research, contributing to the development of AI systems that integrate persuasive elements effectively and ethically.


\section{Ethics Statement} 
\subsection{Data Collection and Privacy} 
The data used in this study necessarily includes personal information such as values and habits. The collected data contains personal information, but anonymization was conducted by assigning unique IDs to each individual. This anonymization helps protect the sensitive personal information of annotators. Additionally, the data was collected through an annotation agency, and during this process, consent was obtained from participants for the use of anonymized personal information. Additionally, appropriate compensation was fairly provided to all participants, reflecting the minimum wage rate and task completion time.

\subsection{Potential Applications and Societal Impact}
This field of study presents the potential to develop technologies capable of creating persuasive images tailored to individuals’ psychological characteristics. Such advancements could meaningfully influence behaviors and decisions, underscoring the need to explore their possible outcomes comprehensively. To ensure constructive applications, future research should focus on identifying approaches that align with ethical principles and societal values.

\subsection{Ethical Responsibilities of Researchers}
The study protocol was reviewed and approved by the Institutional Review Board (IRB), ensuring all procedures complied with ethical standards for research involving human subjects. Additionally, we explicitly acknowledge that this data will not be used for purposes other than this study.

\section{Acknowledgments}
This work was supported by the Creative-Pioneering Researchers Program through Seoul National University. It was also supported by the National Research Foundation of Korea (NRF) grants (RS-2024-00333484, RS-2022 NR070855) and the Institute of Information \& Communications Technology Planning \& Evaluation (IITP) grant [NO.RS-2021-II211343, Artificial Intelligence Graduate School Program (Seoul National University)] funded by the Korean government (MSIT).

\bibliography{custom}

\begin{thebibliography}{37}
\providecommand{\natexlab}[1]{#1}

\bibitem[{Ajzen(1985)}]{ajzen1985intentions}
Icek Ajzen. 1985.
\newblock From intentions to actions: A theory of planned behavior.
\newblock In \emph{Action control: From cognition to behavior}, pages 11--39. Springer.

\bibitem[{Ajzen(1987)}]{ajzen1987attitudes}
Icek Ajzen. 1987.
\newblock Attitudes, traits, and actions: Dispositional prediction of behavior in personality and social psychology.
\newblock In \emph{Advances in experimental social psychology}, volume~20, pages 1--63. Elsevier.

\bibitem[{Ajzen(1991)}]{ajzen1991theory}
Icek Ajzen. 1991.
\newblock The theory of planned behavior.
\newblock \emph{Organizational Behavior and Human Decision Processes}.

\bibitem[{Chandler and Munday(2011)}]{chandler2011dictionary}
Daniel Chandler and Rod Munday. 2011.
\newblock \emph{A dictionary of media and communication}.
\newblock Oxford University Press, USA.

\bibitem[{Dettmers et~al.(2023)Dettmers, Pagnoni, Holtzman, and Zettlemoyer}]{NEURIPS2023_1feb8787}
Tim Dettmers, Artidoro Pagnoni, Ari Holtzman, and Luke Zettlemoyer. 2023.
\newblock \href {https://proceedings.neurips.cc/paper_files/paper/2023/file/1feb87871436031bdc0f2beaa62a049b-Paper-Conference.pdf} {Qlora: Efficient finetuning of quantized llms}.
\newblock In \emph{Advances in Neural Information Processing Systems}, volume~36, pages 10088--10115. Curran Associates, Inc.

\bibitem[{Dimitrov et~al.(2021)Dimitrov, Ali, Shaar, Alam, Silvestri, Firooz, Nakov, and Martino}]{dimitrov2021semeval}
Dimitar Dimitrov, Bishr~Bin Ali, Shaden Shaar, Firoj Alam, Fabrizio Silvestri, Hamed Firooz, Preslav Nakov, and Giovanni Da~San Martino. 2021.
\newblock Semeval-2021 task 6: Detection of persuasion techniques in texts and images.
\newblock \emph{arXiv preprint arXiv:2105.09284}.

\bibitem[{Feinberg and Willer(2019)}]{feinberg2019moral}
Matthew Feinberg and Robb Willer. 2019.
\newblock Moral reframing: A technique for effective and persuasive communication across political divides.
\newblock \emph{Social and Personality Psychology Compass}, 13(12):e12501.

\bibitem[{Garber and Hyatt(2003)}]{garber2003color}
Lawrence~L Garber and Eva~M Hyatt. 2003.
\newblock Color as a tool for visual persuasion.
\newblock In \emph{Persuasive imagery}, pages 313--336. Routledge.

\bibitem[{Goldberg(2013)}]{Big5}
Lewis~R Goldberg. 2013.
\newblock An alternative “description of personality”: The big-five factor structure.
\newblock In \emph{Personality and Personality Disorders}, pages 34--47. Routledge.

\bibitem[{Graham et~al.(2013)Graham, Haidt, Koleva, Motyl, Iyer, Wojcik, and Ditto}]{Moral_foundation_theory}
Jesse Graham, Jonathan Haidt, Sena Koleva, Matt Motyl, Ravi Iyer, Sean~P Wojcik, and Peter~H Ditto. 2013.
\newblock Moral foundations theory: The pragmatic validity of moral pluralism.
\newblock In \emph{Advances in experimental social psychology}, volume~47, pages 55--130. Elsevier.

\bibitem[{Graham et~al.(2011)Graham, Nosek, Haidt, Iyer, Koleva, and Ditto}]{MFQ-30}
Jesse Graham, Brian~A Nosek, Jonathan Haidt, Ravi Iyer, Spassena Koleva, and Peter~H Ditto. 2011.
\newblock Mapping the moral domain.
\newblock \emph{Journal of personality and social psychology}, 101(2):366.

\bibitem[{Hosu et~al.(2020)Hosu, Lin, Sziranyi, and Saupe}]{Hosu.2020}
Vlad Hosu, Hanhe Lin, Tamas Sziranyi, and Dietmar Saupe. 2020.
\newblock \href {https://doi.org/10.1109/tip.2020.2967829} {{KonIQ-10k: An Ecologically Valid Database for Deep Learning of Blind Image Quality Assessment}}.
\newblock \emph{IEEE Transactions on Image Processing}, 29:4041--4056.

\bibitem[{Hussain et~al.(2017)Hussain, Zhang, Zhang, Ye, Thomas, Agha, Ong, and Kovashka}]{hussain2017automatic}
Zaeem Hussain, Mingda Zhang, Xiaozhong Zhang, Keren Ye, Christopher Thomas, Zuha Agha, Nathan Ong, and Adriana Kovashka. 2017.
\newblock Automatic understanding of image and video advertisements.
\newblock In \emph{Proceedings of the IEEE conference on computer vision and pattern recognition}, pages 1705--1715.

\bibitem[{Joo et~al.(2014)Joo, Li, Steen, and Zhu}]{joo2014visual}
Jungseock Joo, Weixin Li, Francis~F Steen, and Song-Chun Zhu. 2014.
\newblock Visual persuasion: Inferring communicative intents of images.
\newblock In \emph{Proceedings of the IEEE conference on computer vision and pattern recognition}, pages 216--223.

\bibitem[{Kong et~al.(2016)Kong, Shen, Lin, Mech, and Fowlkes}]{Kong.2016}
Shu Kong, Xiaohui Shen, Zhe Lin, Radomir Mech, and Charless Fowlkes. 2016.
\newblock \href {https://doi.org/10.48550/arxiv.1606.01621} {{Photo Aesthetics Ranking Network with Attributes and Content Adaptation}}.
\newblock \emph{arXiv}.

\bibitem[{Kumar et~al.(2023)Kumar, Jha, Gupta, Aggarwal, Garg, Malyan, Bhardwaj, Shah, Krishnamurthy, and Chen}]{kumar2023persuasion}
Yaman Kumar, Rajat Jha, Arunim Gupta, Milan Aggarwal, Aditya Garg, Tushar Malyan, Ayush Bhardwaj, Rajiv~Ratn Shah, Balaji Krishnamurthy, and Changyou Chen. 2023.
\newblock Persuasion strategies in advertisements.
\newblock In \emph{Proceedings of the AAAI conference on artificial intelligence}, volume~37, pages 57--66.

\bibitem[{Liu et~al.(2022)Liu, Guo, Dai, and Litman}]{liu2022imagearg}
Zhexiong Liu, Meiqi Guo, Yue Dai, and Diane Litman. 2022.
\newblock \href {https://aclanthology.org/2022.argmining-1.1} {{I}mage{A}rg: A multi-modal tweet dataset for image persuasiveness mining}.
\newblock In \emph{Proceedings of the 9th Workshop on Argument Mining}, pages 1--18, Online and in Gyeongju, Republic of Korea. International Conference on Computational Linguistics.

\bibitem[{Messaris(1996)}]{messaris1996visual}
Paul Messaris. 1996.
\newblock \emph{Visual persuasion: The role of images in advertising}.
\newblock sage publications.

\bibitem[{Miller(1998)}]{miller1998visual}
Thomas Miller. 1998.
\newblock Visual persuasion: A comparison of visuals in academic texts and the popular press.
\newblock \emph{English for specific purposes}, 17(1):29--46.

\bibitem[{Nordmo and Selart(2015)}]{mads_nordmo__2015}
Mads Nordmo and Marcus Selart. 2015.
\newblock \href {https://doi.org/10.3389/FPSYG.2015.01324} {The asymmetrical force of persuasive knowledge across the positive-negative divide.}
\newblock \emph{Frontiers in Psychology}.

\bibitem[{Orji et~al.(2016)Orji, Busch, Dijkstra, Reisinger, Stibe, and Tscheligi}]{orji2016personalization}
Rita Orji, Marc Busch, Arie Dijkstra, Michaela Reisinger, Agnis Stibe, and Manfred Tscheligi. 2016.
\newblock Personalization in persuasive technology.
\newblock In \emph{Adjunct Proceedings of the 11th International Conference on Persuasive Technology}, pages 96--99.

\bibitem[{Oyibo et~al.(2017)Oyibo, Orji, and Vassileva}]{oyibo2017investigation}
Kiemute Oyibo, Rita Orji, and Julita Vassileva. 2017.
\newblock Investigation of the influence of personality traits on cialdini's persuasive strategies.
\newblock \emph{PPT@ PERSUASIVE}, 2017:8--20.

\bibitem[{Park et~al.(2014)Park, Shim, Chatterjee, Sagae, and Morency}]{park2014computational}
Sunghyun Park, Han~Suk Shim, Moitreya Chatterjee, Kenji Sagae, and Louis-Philippe Morency. 2014.
\newblock Computational analysis of persuasiveness in social multimedia: A novel dataset and multimodal prediction approach.
\newblock In \emph{Proceedings of the 16th international conference on multimodal interaction}, pages 50--57.

\bibitem[{Rammstedt and John(2007)}]{rammstedt2007measuring}
Beatrice Rammstedt and Oliver~P John. 2007.
\newblock Measuring personality in one minute or less: A 10-item short version of the big five inventory in english and german.
\newblock \emph{Journal of research in Personality}, 41(1):203--212.

\bibitem[{Ren et~al.(2017)Ren, Shen, Lin, Měch, and Foran}]{Ren.2017}
Jian Ren, Xiaohui Shen, Zhe Lin, Radomír Měch, and David~J. Foran. 2017.
\newblock \href {https://doi.org/10.1109/iccv.2017.76} {{Personalized Image Aesthetics}}.
\newblock \emph{2017 IEEE International Conference on Computer Vision (ICCV)}, pages 638--647.

\bibitem[{Ruiz-Arellano et~al.(2022)Ruiz-Arellano, Mejía-Medina, Castillo-Topete, Fong-Mata, Hernández-Torres, Rodríguez-Valenzuela, and Berra-Ruiz}]{AI_visual_persuasive}
Ana~Erika Ruiz-Arellano, David~Abdel Mejía-Medina, Victor~H. Castillo-Topete, María~Berenice Fong-Mata, Ervey~L. Hernández-Torres, Paloma Rodríguez-Valenzuela, and Enrique Berra-Ruiz. 2022.
\newblock \href {https://doi.org/10.3390/designs6060124} {Addressing the use of artificial intelligence tools in the design of visual persuasive discourses}.
\newblock \emph{Designs}, 6(6):124--124.

\bibitem[{Schwartz(2006)}]{schwartz2006theory}
Shalom Schwartz. 2006.
\newblock A theory of cultural value orientations: Explication and applications.
\newblock \emph{Comparative sociology}, 5(2-3):137--182.

\bibitem[{Schwartz(2012)}]{schwartz2012overview}
Shalom~H Schwartz. 2012.
\newblock An overview of the schwartz theory of basic values.
\newblock \emph{Online readings in Psychology and Culture}, 2(1):11.

\bibitem[{Schwartz et~al.(2012)Schwartz, Cieciuch, Vecchione, Davidov, Fischer, Beierlein, Ramos, Verkasalo, L{\"o}nnqvist, Demirutku et~al.}]{schwartz2012refining}
Shalom~H Schwartz, Jan Cieciuch, Michele Vecchione, Eldad Davidov, Ronald Fischer, Constanze Beierlein, Alice Ramos, Markku Verkasalo, Jan-Erik L{\"o}nnqvist, Kursad Demirutku, et~al. 2012.
\newblock Refining the theory of basic individual values.
\newblock \emph{Journal of personality and social psychology}, 103(4):663.

\bibitem[{Seo(2020)}]{seo2020meta}
Kiwon Seo. 2020.
\newblock Meta-analysis on visual persuasion--does adding images to texts influence persuasion.
\newblock \emph{Athens Journal of Mass Media and Communications}, 6(3):177--190.

\bibitem[{Singh et~al.(2024)Singh, Singla, SI, and Krishnamurthy}]{singh2024measuring}
Somesh Singh, Yaman~K Singla, Harini SI, and Balaji Krishnamurthy. 2024.
\newblock Measuring and improving persuasiveness of large language models.
\newblock \emph{arXiv preprint arXiv:2410.02653}.

\bibitem[{Tversky and Kahneman(1981)}]{tversky1981framing}
Amos Tversky and Daniel Kahneman. 1981.
\newblock The framing of decisions and the psychology of choice.
\newblock \emph{science}, 211(4481):453--458.

\bibitem[{Voelkel and Feinberg(2018)}]{voelkel2018morally}
Jan~G Voelkel and Matthew Feinberg. 2018.
\newblock Morally reframed arguments can affect support for political candidates.
\newblock \emph{Social Psychological and Personality Science}, 9(8):917--924.

\bibitem[{Walton et~al.(2008)Walton, Reed, and Macagno}]{walton2008argumentation}
Douglas Walton, Christopher Reed, and Fabrizio Macagno. 2008.
\newblock \emph{Argumentation schemes}.
\newblock Cambridge University Press.

\bibitem[{Wang et~al.(2019)Wang, Shi, Kim, Oh, Yang, Zhang, and Yu}]{wang2019persuasion}
Xuewei Wang, Weiyan Shi, Richard Kim, Yoojung Oh, Sijia Yang, Jingwen Zhang, and Zhou Yu. 2019.
\newblock Persuasion for good: Towards a personalized persuasive dialogue system for social good.
\newblock \emph{arXiv preprint arXiv:1906.06725}.

\bibitem[{Webb and Sheeran(2006)}]{webb2006does}
Thomas~L Webb and Paschal Sheeran. 2006.
\newblock Does changing behavioral intentions engender behavior change? a meta-analysis of the experimental evidence.
\newblock \emph{Psychological bulletin}, 132(2):249.

\bibitem[{Yang et~al.(2022)Yang, Xu, Li, Qie, Li, Zhang, and Guo}]{Yang.20220d}
Yuzhe Yang, Liwu Xu, Leida Li, Nan Qie, Yaqian Li, Peng Zhang, and Yandong Guo. 2022.
\newblock \href {https://doi.org/10.1109/cvpr52688.2022.01924} {{Personalized Image Aesthetics Assessment with Rich Attributes}}.
\newblock \emph{2022 IEEE/CVF Conference on Computer Vision and Pattern Recognition (CVPR)}, 00:19829--19837.

\end{thebibliography}

\clearpage
\appendix

\section{PVP Dataset }\label{sec:Dataset}

\subsection{Topics and Messages }\label{subsubsec:Data Construction_Messages}

\label{app:Messages}

To collect a variety of messages, the 15 executive departments of the United States were referenced, and 5 additional messages were created to represent a broader range of topics. The topics, descriptions, and examples can be found in Table~\ref{tab:topics}.

After selecting the topics, messages were generated using GPT-4 with the Prompt~\ref{lst: prompt for generating messages}, while varying the designation of the topic.

\begin{table*}[hbt!]
\centering
\small
\begin{tabularx}{\linewidth}{@{}p{2cm}p{8cm}X@{}}    \toprule
    Topic & Description & Example Messages \\ \midrule 
    Agriculture & Sustainable food choices, food safety, and eco-friendly practices. & Purchase organic food, Consume seasonal produce\\
    Commerce & Staying informed about economic trends and business knowledge. & Read economic newspapers, Study consumer rights \\
    Defense & Understanding military operations and national security. & Visit military bases, Prepare for national emergencies \\
    Education & Creating optimal study environments and healthy habits. & Do not listen to music while studying, Do not watch TV while studying\\
    Energy & Conserving energy through efficient practices and devices. & Use energy-efficient products, Turn off unnecessary lights and use natural light\\
    Health and Human Services & Maintaining a healthy lifestyle with balanced nutrition and exercise. & Reduce sugar intake,Eat seasonal foods  \\
    Homeland Security & Personal safety, online security, and disaster preparedness.& Do not cross the border, Do not visit dangerous countries\\
    Housing and Urban Development & Supporting housing initiatives and maintaining home safety and efficiency. & Choose energy-efficient homes, Get home insurance \\
    Interior & Preserving nature and practicing eco-friendly habits.& Take your trash with you after a picnic, Pick up trash at the beach\\
    Labor & Adhering to safety protocols and using protective equipment. & Wear a safety helmet, Wear work clothes \\
    State & Engaging with and respecting diverse cultures.& Try multicultural foods, Enjoy music from different cultures\\
    Transportation & Practicing safe driving and pedestrian safety measures.& Fasten your seatbelt, Reduce your speed when there are many pedestrians \\
    Treasury & Adopting frugal habits and smart financial practices. & Buy clothes during sales, Set a savings goal each month\\
    Veterans Affairs (VA) & Expressing gratitude and recognizing military personnel's efforts. & Thank soldiers, Participate in military events\\ \midrule
    Safety Awareness & Maintaining safety and caution in daily activities. &Be careful when using a knife, Be careful when touching a hot pot\\
    Exercise & Incorporating regular physical activity and exercise. & Do yoga every morning, Stretch before going to bed\\
    Privacy & Protecting personal privacy and security in all environments. & Use a privacy screen on your phone, Do not share personal information on social media\\
    Cyber Etiquette & Maintaining respectful and responsible online behavior. & Do not post malicious comments, Do not use foul language in games\\
    Advertising & Adopting stylish and trendy products. & Use an iPhone instead of a Galaxy to look younger, Use a MacBook in cafes to look cool\\ \bottomrule
\end{tabularx}
\caption{Topics and their descriptions, with example messages for each topic. The first 15 topics correspond to the executive departments of the U.S. government.}
\label{tab:topics}
\end{table*}

\begin{lstlisting}[caption={Prompt 1: Prompt for generating messages}, label={lst: prompt for generating messages}]
Please create persuasive messages that demand behavioral change, following these conditions:

    1.	They must be universal and not violate common sense.
    2.	They must be immediately relatable and something that an average person can do.
    3.	The topic should be about {description, e.g., Sustainable food choices, food safety, and eco-friendly practices}.
    4.	Generate 30 distinct messages that do not overlap with each other. 
    5. Exclude any reasoning; the messages should be direct and action-oriented.

Here is an example:
    1.	Purchase organic food.
    2.	Consume seasonal produce.



\end{lstlisting}

\subsection{Premises} \label{sec:Data Construction_Premises}
\label{app:Pemise}

We generate premises to reflect the message and the intended persuasion strategy. Positive framing focuses on the desired behavior, such as `Doing exercise everyday', whereas negative framing emphasizes the opposite behavior, such as `Not doing exercise everyday'. Additionally, the prompt is written in the progressive tense. We use the following prompt to generate premises.

\begin{lstlisting}[caption={Prompt 2: Prompt for generating premises}, label={lst: prompt for generating premises}]
{Script 1}
each item should satisfy the following criteria: 
- give 6 phrases 
- do not generate explanation 
- generate concrete and succinct phrase 
- provide phrases in the following keys: "1", "2", "3", "4", "5", "6"
{Script 2}
\end{lstlisting}
The content of Script 1 and Script 2 varies depending on the persuasion strategy, as shown in Table~\ref{tab:premise_piping}.

\begin{table*}[hbt!]
\centering
\small
\begin{tabularx}{\linewidth}{@{}lXp{0.35\linewidth}@{}}    
    \toprule
    \textbf{Strategy} & \textbf{Script 1} & \textbf{Script 2} \\ 
    \midrule 
    Positive Perceived Persona & Provide a brief list of positive persona or attribute as perceived by others of \{message\}. & Please generate the script while keeping the form `Positive effect: You can be seen as \_\_\_.' \\
    
    Negative Perceived Persona & Provide a brief list of negative persona or attribute as perceived by others of \{message\}. & Please generate the script while keeping the form `Negative effect: You can be seen as \_\_\_.' \\
    
    Positive Internal Emotion & Provide a brief list of the concrete, positive, and emotional reactions you have when you do this message: \{message\}. & Please generate the script while keeping the form `Positive effect: You can feel \_\_\_.' \\
    
    Negative Internal Emotion & Provide a brief list of the concrete, negative, and emotional reactions you have when you do this message: \{message\}. & Please generate the script while keeping the form `Negative effect: You can feel \_\_\_.' \\
    
    Positive External Emotion & I want to obtain the results for the following script: When you take the following action for yourself, here are the concrete, positive, and emotional responses others feel towards you in that situation: \{message\}. & Please generate the script while keeping the form `Positive effect: Others can \_\_\_.' \\
    
    Negative External Emotion & I want to obtain the results for the following script: When you take the following action for yourself, here are the concrete, negative, and emotional responses others feel towards you in that situation: \{message\}. & Please generate the script while keeping the form `Negative effect: Others can \_\_\_.' \\
    
    Positive Consequence & Provide a brief list of positive and concrete consequences of \{message\}. & Please generate the script while keeping the form `Positive effect:' \\
    
    Negative Consequence & Provide a brief list of negative and concrete consequences of \{message\}. & Please generate the script while keeping the form `Negative effect:' \\
    
    Bandwagon & Provide a brief list suggesting that something should be accepted because it is popular or everyone is doing it for \{message\}. & None \\
    
    \bottomrule
\end{tabularx}
\caption{Scripts used to generate premises reflecting positive and negative framing strategies.}
\label{tab:premise_piping}
\end{table*}

\subsection{Queries}\label{sec:Data Construction_Queries}
\label{app:Queries}
\subsubsection{Optimizing DALLE Prompts for Image Generation}
DALLE prompts are designed to provide clear and specific instructions for image generation. To create an effective prompt, it’s crucial to describe the key elements of the image in detail, ensuring that they can be visually represented. This involves giving precise directions on aspects such as color, composition, mood, expressions, gestures, and background, so that each element strengthens the intended message. Additionally, the main theme of the image should be visually emphasized, with supplementary elements supporting it. 
Lastly, specifying the size and placement of each element is essential to ensure that the focal point aligns with the theme. Following these guidelines will result in images that effectively communicate the intended message.

\begin{lstlisting}[caption={Prompt 3: Prompt for generating a query for DALLE}, label={lst: prompt for Converting Query_Dalle}]
I want to generate an image based on the theme: {premise} and the message: {message}. 

Please describe in detail how to represent this theme in an image.

The result should be phrased as a complete sentence, emphasize the theme, and not exceed 10 sentences.
\end{lstlisting}

\subsubsection{Optimizing Google Image Search Queries}
In developing a complex script for Google Image Search, search queries in sentence form were not effective in yielding images that accurately reflected the intended meaning. Instead, they often produced overly specific or irrelevant results. To address this issue, we chose to use short phrases, instead. By crafting queries that are concise and limited to five words or fewer, we managed to obtain the search results that were more accurate and relevant. This approach ensures that the search terms are both succinct and focused, thereby enhancing the precision and relevance of the images retrieved.



\begin{lstlisting}[caption={Prompt 4: Prompt for generating a query for Google Search Image}, label={lst: prompt for Converting Query_Google_word}]
I need a single search query for Google Images based on the description: {premise} and the goal: {message}. 
The query should meet these criteria:
    1.	Be concrete and succinct
    2.	Contain no more than 5 words
    3.	Be formatted in a list with commas separating words
    4.	No explanations, just the query
    5.	No quotation marks.

\end{lstlisting}

\subsection{Images} \label{app:Images}
\label{sec:Data Construction_Images}

In this section, we provide a detailed explanation of the process for obtaining and validating images. Using the methods described in the previous sections, we first generate 6 premises and their corresponding queries for each persuasion strategy. Then, we use DALLE-3 and Google Image Search to obtain images. For each image source (DALLE and Google), our goal is to acquire three images per persuasion strategy.

\subsubsection{DALLE}
For each query, we generate a single image with the following prompt. Then the image goes through the validation process (described below). Among the images generated from the 6 queries, the first 3 images that pass the validation process are used in the dataset.
\begin{lstlisting}[caption={Prompt 5: Prompt for generating images (DALLE)}, label={lst: prompt for generating images, dalle}]
Generate image: {query}. The generated image should be created in a photorealistic style.
\end{lstlisting}

\subsubsection{Google}
For each query, we crawled 40 images from Google Image Search using the Serper API. To filter out text-heavy images, we applied OCR using the easyocr library and excluded images with more than 20 characters (with OCR confidence over 0.95). Then, the images went through the validation process (described below), and the first 3 images to pass the validation process were included in the dataset.

\subsubsection{Validation}
We used GPT-4o to validate whether the images effectively convey the intended premise. We conducted experiments with various prompts and would like to share insights from our trials and errors. 
In our initial design, we provided GPT with an image and the intended premise, asking GPT to rate how well the image reflects the premise on a scale from 0 to 10. However, we observed that GPT tends to award scores too high. Even when GPT recognizes that the image does not reflect the premise (via chain-of-thought), it avoided giving a low score. 

To mitigate this, we changed the evaluation process by explicitly dividing it into two steps: (1) Asking GPT to interpret the image, and (2) scoring the interpretation as to whether it aligns with the premise. This allows GPT to focus on whether the initial impression of the image reflects the premise. The following prompts are used for steps (1) and (2). Note that for step (2), different prompts were used for different persuasion strategies.

While there is a protocol in place to remove offensive content when generating images using GPT prompts, ensuring that uncomfortable or harmful material is filtered out, we still went through a manual process to further remove any offensive content. This involved eliminating violent images, expressions that could be offensive to individuals, and content that could cause sexual embarrassment or discomfort.

\begin{lstlisting}[caption={Prompt 6: Prompt for validation step (1)}, label={lst: prompt for evaluating images, first turn}]
Attached is an image about {message}. What message does this image intend to convey?
\end{lstlisting}

\begin{lstlisting}[caption={Prompt 7: Prompt for step (2) (Perceived Persona)}, label={lst: prompt for evaluating images, second turn, perceived persona}]
You are a helpful assistant designed to output JSON.

The actual message that the image intended to convey is {premise}.

How well does your interpretation capture the persona or attributes of the person who conducts this action perceived by other people as described in the intended message?

Give a brief explanation in the "reason" key. Rate the score between 0 and 10 (0: not captured at all, 10: perfectly captured). Provide your rating in the "score" key.
\end{lstlisting}

\begin{lstlisting}[caption={Prompt 8: Prompt for step (2) (Internal Emotion)}, label={lst: prompt for evaluating images, second turn, internal emotion}]
You are a helpful assistant designed to output JSON.

The actual message that the image intended to convey is {premise}.

How well does your interpretation capture the emotional reactions of the person who conducts this action as described in the intended message?
    
Give a brief explanation in the "reason" key. Rate the score between 0 and 10 (0: not captured at all, 10: perfectly captured). Provide your rating in the "score" key.
\end{lstlisting}

\begin{lstlisting}[caption={Prompt 9: Prompt for step (2) (External Emotion)}, label={lst: prompt for evaluating images, second turn, external emotion}]
You are a helpful assistant designed to output JSON.

The actual message that the image intended to convey is {premise}.

How well does your interpretation capture the feelings that this action may cause to other people as described in the intended message?

Give a brief explanation in the "reason" key. Rate the score between 0 and 10 (0: not captured at all, 10: perfectly captured). Provide your rating in the "score" key.
\end{lstlisting}

\begin{lstlisting}[caption={Prompt 10: Prompt for step (2) (Consequence)}, label={lst: prompt for evaluating images, second turn, consequence}]
You are a helpful assistant designed to output JSON.

The actual message that the image intended to convey is {premise}.

How well does your interpretation capture the consequences of this action described in the intended message?

Give a brief explanation in the "reason" key. Rate the score between 0 and 10 (0: not captured at all, 10: perfectly captured). Provide your rating in the "score" key.
\end{lstlisting}

\begin{lstlisting}[caption={Prompt 11: Prompt for step (2) (Bandwagon)}, label={lst: prompt for evaluating images, second turn, bandwagon}]
You are a helpful assistant designed to output JSON.

The actual message that the image intended to convey is {premise}.

How well your interpretation capture the collective behavior of many people or popular opinion in the intended message?
    
Give a brief explanation in the "reason" key. Rate the score between 0 and 10 (0: not captured at all, 10: perfectly captured). Provide your rating in the "score" key.
\end{lstlisting}

While \textbf{using GPT as an evaluator} has been widely adopted in recent research, it is important to validate whether GPT's scores align with human judgments. Therefore, we selected 50 images (25 from Google and 25 from DALLE) and compared GPT's scoring with that of human evaluators (two co-authors). 
It is important to take into account the variability of GPT scoring and choose the optimal number of scoring for each image. To that end, we first scored each image 40 times using GPT-4o and looked for the optimal number using bootstrapping. Specifically, from these 40 scores, we randomly selected \textit{N} scores and calculated the correlation between the average of the sampled scores and the scores given by humans. We repeated this process 1,000 times, generating a distribution of correlations (i.e., bootstrapping). 
As we increased \textit{N}, the 95\% confidence interval narrowed, decreasing the variability in correlation.
However, it also increased cost as we need more GPT API calls, creating a trade-off between evaluation reliability and cost. Based on the above bootstrapping analysis, we decided to use the following criteria to choose \textit{N}: (1) a lower bound of the confidence interval greater than 0.3, and (2) a correlation coefficient of at least 0.5. Consequently, we decided to set \textit{N} as 3. For reference, the validation correlation for DALLE was 0.501, and 0.557 for Google.




\subsection{Annotation Validation}\label{app:Annotation Validation} \label{sec:Data Construction_Score}

For each image, an annotator was asked to answer the following question: ``Assuming you do not engage in a specific behavior, please rate on a scale from 0 to 10 how much you feel promoted to perform that behavior after viewing each image.'' The annotator also completed three questionnaires about psychological characteristics (BFI-10, PVQ-21, MFQ-30).

To filter out unreliable annotations, we used the following process:
\begin{itemize}
    \item~\textbf{Response Variance Criterion:} Annotations where the variance of persuasiveness scores across images was below 0.1 were considered unreliable. This threshold was chosen in consideration of variances resulting from random scoring (e.g., choosing a score of 10 for all images except one). Based on our pilot study, we found 0.1 to be an appropriate threshold.
    \item~\textbf{Duplicate Image Evaluation:} To ensure that annotators stayed focused on the task, we included duplicates of three images for each message. These images were carefully chosen based on their quality (either very high or very low) so that they would likely receive consistent persuasiveness scores. If the score difference between any duplicates exceeded 2 points, the annotation was considered unreliable.
    \item~\textbf{Big Five Questionnaire Evaluation:} The BFI-10 questionnaire for the Big Five includes two questions for each personality dimension. We measured the internal consistency of responses, while reversing the scores for reverse-scored items. If the score difference between two questions for the same personality dimension was above 2, then the annotation was considered unreliable. This method was not applied to other questionnaires, as similar items might yield different scores depending on their content.
\end{itemize}
Three annotators failed to meet two or more of these criteria, and their annotations were excluded from our dataset.

\subsection{Inter-Annotator Agreement}
We measured inter-annotator agreement using Fleiss' Kappa score under two different settings (Table~\ref{tab:top5_bottom5_inter_annotator}).

In the first setting, each distinct persuasiveness score was treated as a separate category. The results presented an average Fleiss' kappa score of -0.027, with a maximum of 0.1 and a minimum of -0.14 depending on the message. These results suggest little to no agreement among annotators.

In the second setting, the scores were re-categorized into broader ranges to account for variability among annotators. Scores from 0 to 2 were grouped into Category 1, scores from 3 to 6 into Category 2, and scores from 7 to 10 into Category 3. Under this categorization, the results showed an average Fleiss' kappa score of -0.030, with maximum of 0.37 and a minimum of -0.23 depending on the message. Even with this re-categorization, the average score remained close to 0, indicating no clear evidence for agreement among annotators.

Interestingly, in the second setting, messages such as ``Do not visit dangerous countries'' or ``Do not swim in deep water'' showed relatively high inter-annotator agreement. This suggests that messages grounded in widely accepted social norms or invoking instinctive safety concerns lead to more consistent judgments across annotators. In contrast, messages like ``Tell ROTC students they look cool'' or ``Save on food expenses'' showed low inter-annotator agreement, indicating that messages based on personal values or with ambiguous validity or relevance are interpreted less consistently.

Our dataset captures a wide range of persuasive strategies targeting diverse human values, which naturally elicit varied responses from annotators with different value profiles. Thus, the observed pluralism in judgements is not only expected but also central to the contribution of our work.

\begin{table}[t]
\resizebox{\columnwidth}{!}{%
\begin{tabular}{ll}
\toprule
Message                                           & Score    \\ \hline
Do not visit dangerous countries & 0.37 \\
Post pictures with pets on social media to look kind                & 0.28 \\
Do not swim in deep water                     & 0.25 \\
Keep children away from the kitchen and guide them to a safe area                  & 0.23 \\ \midrule
Tell ROTC students they look cool & -0.23 \\
Install solar panels               & -0.22 \\
Install solar panels for home use                     & -0.21 \\
Save on food expenses                 & -0.21 \\ \bottomrule
\end{tabular}%
}
\caption{Top four and bottom four messages in terms of inter annotator agreement in second setting.}  
\label{tab:top5_bottom5_inter_annotator}
\end{table}

\section{Data Analysis}




\subsection{Personality and Values} \label{app:Personality and Values}
We first categorized the personality or value scores into four bins, ranging from 1 to 4. After that, for every pair of strategy and personality trait or value, we calculated the Spearman correlation between the persuasiveness scores of images associated with that strategy and the binned personality/value scores of the annotators who rated those images. A high correlation indicates that a particular persuasion strategy is more effective as individuals possess certain personality traits or hold values more strongly.

\subsection{Correlations between Personality/Values and Topics}\label{sec:correlation_personality_topics}
\label{app:Correlations between Personality/Values and Topics}
\begin{figure*}[hbt!]
    \centering
    \includegraphics[width=1\linewidth]{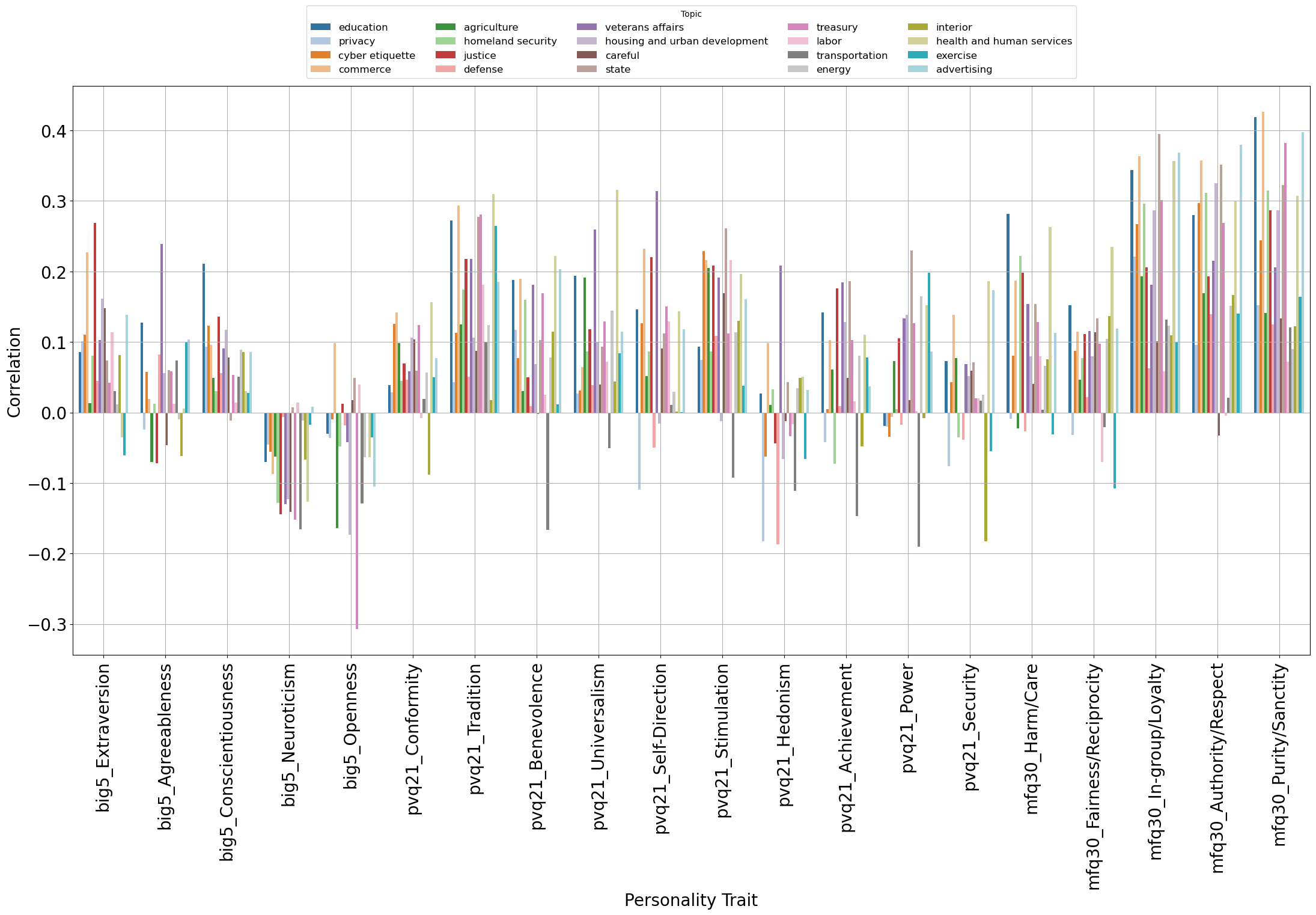}
    \includegraphics[width=1\linewidth]{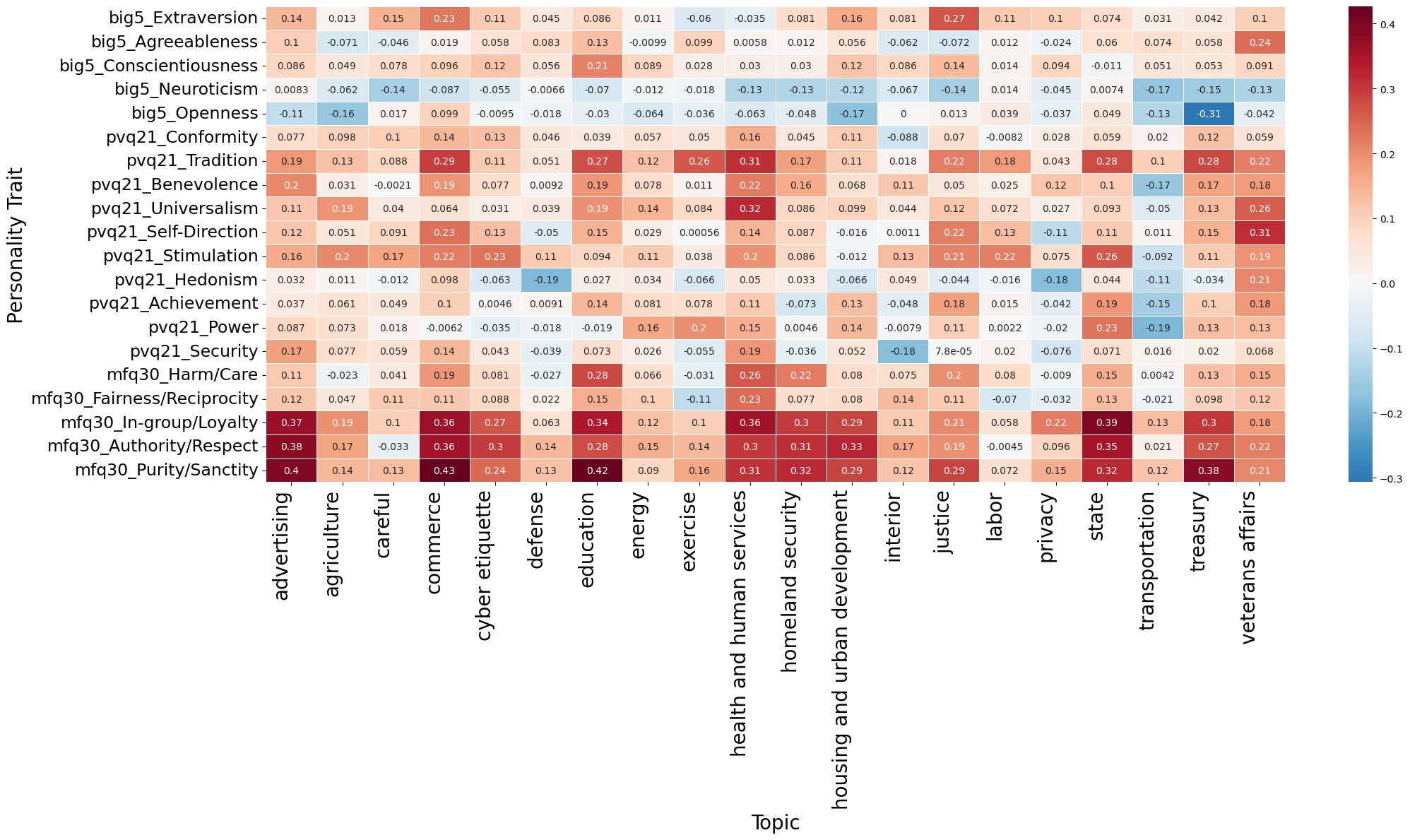}
    \caption{Correlation between psychological characteristics and persuasiveness across various topics. These plots display the correlation coefficients between different psychological characteristics (as measured by MFQ-30) and the persuasiveness of messages across a range of topics.}
    \label{fig:correlation_topic}
\end{figure*}

Several moral values exhibit significant and consistent correlations across topics (Figure ~\ref{fig:correlation_topic}). Topics like
advertising, commerce, health and human services, homeland security, and state show significant correlations (all above 0.30) with multiple moral foundations, including In-group/Loyalty, Authority/Respect, and Purity/Sanctity.

Topics like advertising and commerce are primarily linked to concerns about adherence to social expectations (In-group/Loyalty), social reputation (Authority/Respect) and moral propriety (Purity/Sanctity). For those who value loyalty and appropriate social behavior, these topics are particularly persuasive.

Similarly, topics like health and human services, homeland security, and state show comparable moral alignments, but in broader and more public contexts. The topic of health and human services emphasizes protection and welfare, homeland security relates to civic duty and institutional trust, and state involves
multicultural messages. For those who seek unity (In-group/Loyalty), shared responsibility (Authority/Respect), and moral coherence (Purity/Sanctity) within diverse communities, these topics are effective.

Additionally, the topic of health and human services show meaningful correlations with Tradition (0.31) and Universalism (0.32). Since this topic emphasizes inclusiveness and empathy, it is effective for individuals who
adhere to established traditions (Tradition) and value concern for others (Universalism).

\subsection{Score Distribution by Psychological Characteristics}
\begin{figure*}[h]
    \centering
    \includegraphics[width=1\linewidth]{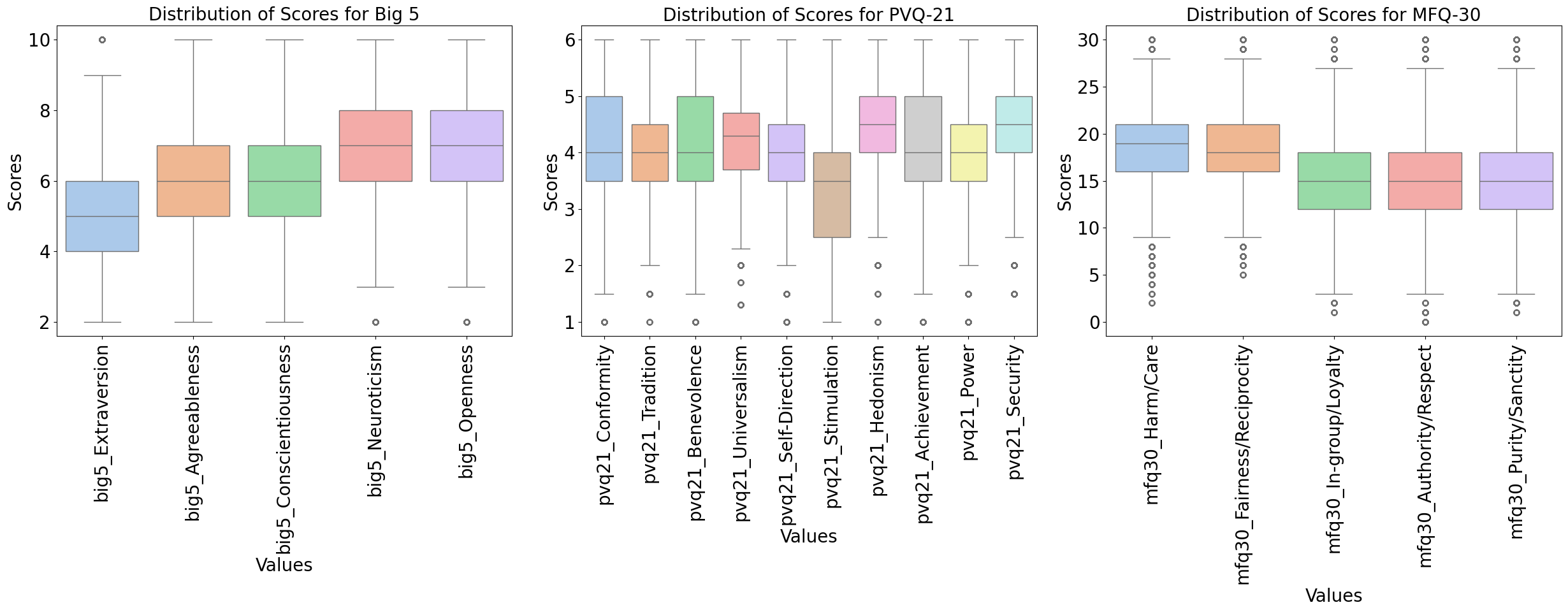}
    \caption{Score Distribution by Personality Traits and Values: Box plots illustrating the range and median scores for personality traits and values based on the Big 5, PVQ-21, and MFQ-30 scales.}
    \label{fig: box_plot-value}
\end{figure*}

In this section, we compared and analyzed the score distributions of respondents using three major scales: BFI-10 (Big Five Inventory), Portrait Values Questionnaire (PVQ-21), and Moral Foundations Questionnaire (MFQ-30) (Figure ~\ref{fig: box_plot-value}). 

This allowed us to understand how respondents’ reactions vary depending on the traits or values addressed by each scale. 

Firstly, in the case of the BFI-10, significant differences in distribution were observed between the traits. For example, Extraversion had a widely dispersed score distribution, while Conscientiousness showed a more concentrated distribution. This suggests that traits like Extraversion can be interpreted quite differently by respondents, indicating notable individual differences.

Secondly, the PVQ-21 values scale also exhibited substantial differences in score distribution depending on the value. Certain values, such as Power, were concentrated in lower score ranges, whereas others, like Achievement and Self-Direction, were more broadly distributed. These differences reflect the possibility that each value may be interpreted differently by respondents, with evaluations potentially varying significantly depending on personal background and experience.

On the other hand, in the case of the MFQ-30, the score distributions were relatively similar across all foundations. There was little variation in distribution between the moral foundations, and the number of outliers was also fairly consistent.  This suggests that moral judgments may be made according to more commonly shared standards among respondents. 
 
\subsection{DALLE vs. Google Image Search} \label{app:DALLE vs. Google Image}
A closer examination shows that obtaining images that accurately reflect intended premises is much easier with DALLE than Google. During the validation phase of our dataset construction, 133,556 images from Google were discarded for they did not accurately represent the intended premises, as opposed to only 1,931 images from DALLE. While Google Image Search often yields more authentic photographs, DALLE-generated images align more closely with specific visual requirements and are perceived as more persuasive. This efficiency and effectiveness highlight the promise of image generation models as valuable tools in crafting personalized visual persuasion strategies.
\begin{table}[t]
    \small
    \begin{tabularx}{\linewidth}{Xl Xl}
        \toprule
        Topic & Score & Topic & Score \\
        \midrule
        transportation & 5.44 & housing and urban development & 4.63 \\
        interior & 5.40 & education & 4.55 \\
        homeland security & 5.29 & health and human services & 4.45 \\
        safety awareness & 5.03 & state & 4.35 \\
        agriculture & 5.02 & commerce & 4.26 \\
        labor & 5.00 & advertising & 4.23 \\
        energy & 4.85 & exercise & 4.19 \\
        justice & 4.85 & cyber etiquette & 4.15 \\
        veterans affairs & 4.81 & treasury & 3.96 \\ 
        privacy & 4.65 & defense & 3.73 \\
        \bottomrule
    \end{tabularx}
    \caption{Average persuasiveness scores by topic (0: Not motivated at all, 10: Highly motivated).}
    \label{tab:average_persuasiveness_scores_topic}
\end{table}

\begin{table}[t]
\resizebox{\columnwidth}{!}{%
\begin{tabular}{ll}
\toprule
Message                                           & Score    \\ \hline
Reduce your speed when there are many pedestrians & 8.19 \\
Go indoors during lightning storms                & 7.80 \\
Clean up trash after fishing                      & 7.75 \\
Do not get into a stranger's car                  & 7.65 \\
Do not cut trees carelessly                       & 7.43 \\ \midrule

Do squats & 1.70 \\
Read books on defense & 1.83 \\
Look up videos of military dog training & 1.87 \\
Learn the differences between the Army, Navy, and Air Force & 1.99 \\
Avoid using earphones while walking & 2.00 \\ \bottomrule

\end{tabular}%
}
\caption{Top five and bottom five messages in terms of average persuasiveness scores.}  
\label{tab:top5_bottom5_messages}
\end{table}


\begin{figure*}[h]
    \centering
    \includegraphics[width=1\linewidth]{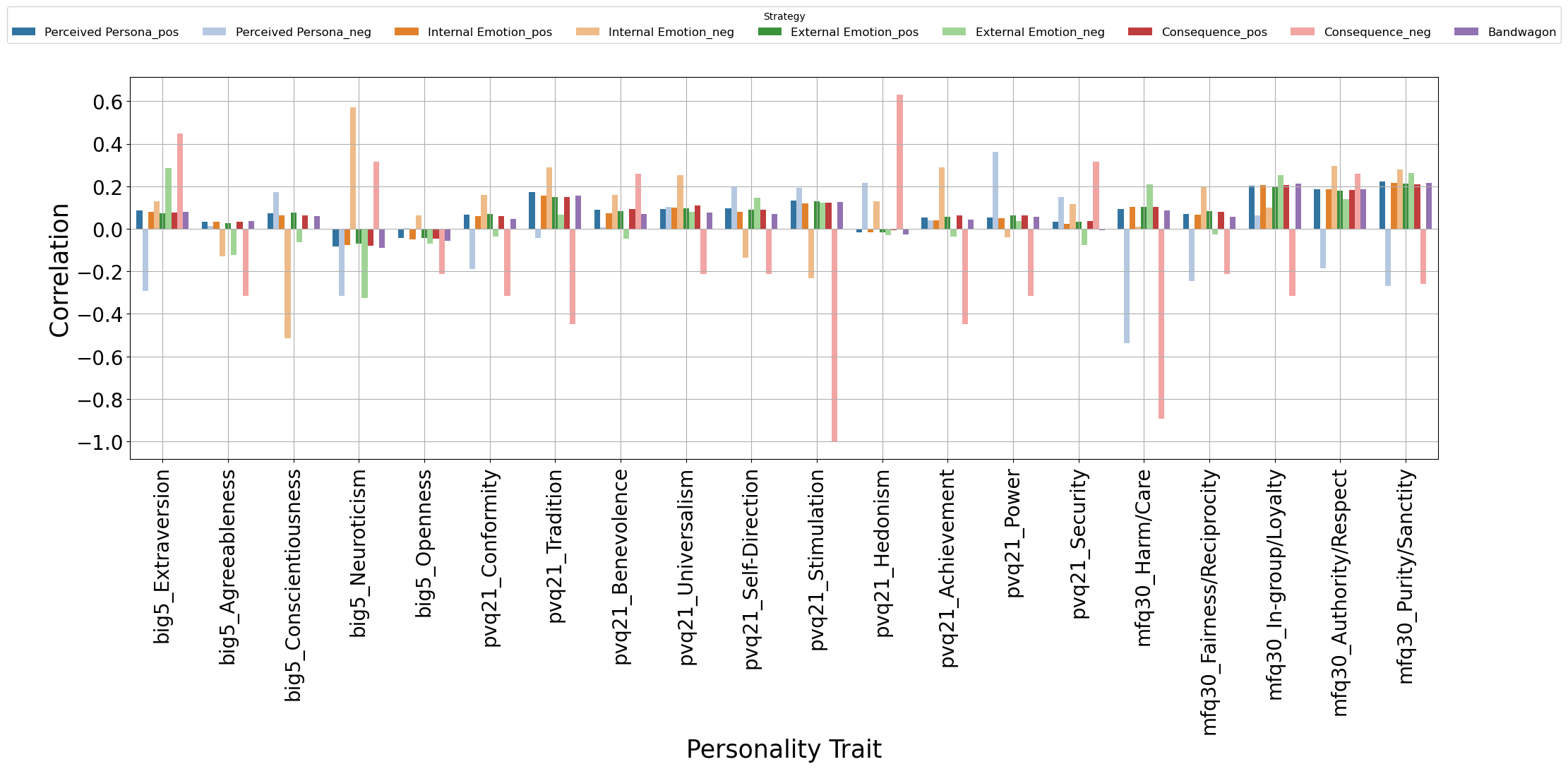}
    \includegraphics[width=1\linewidth]{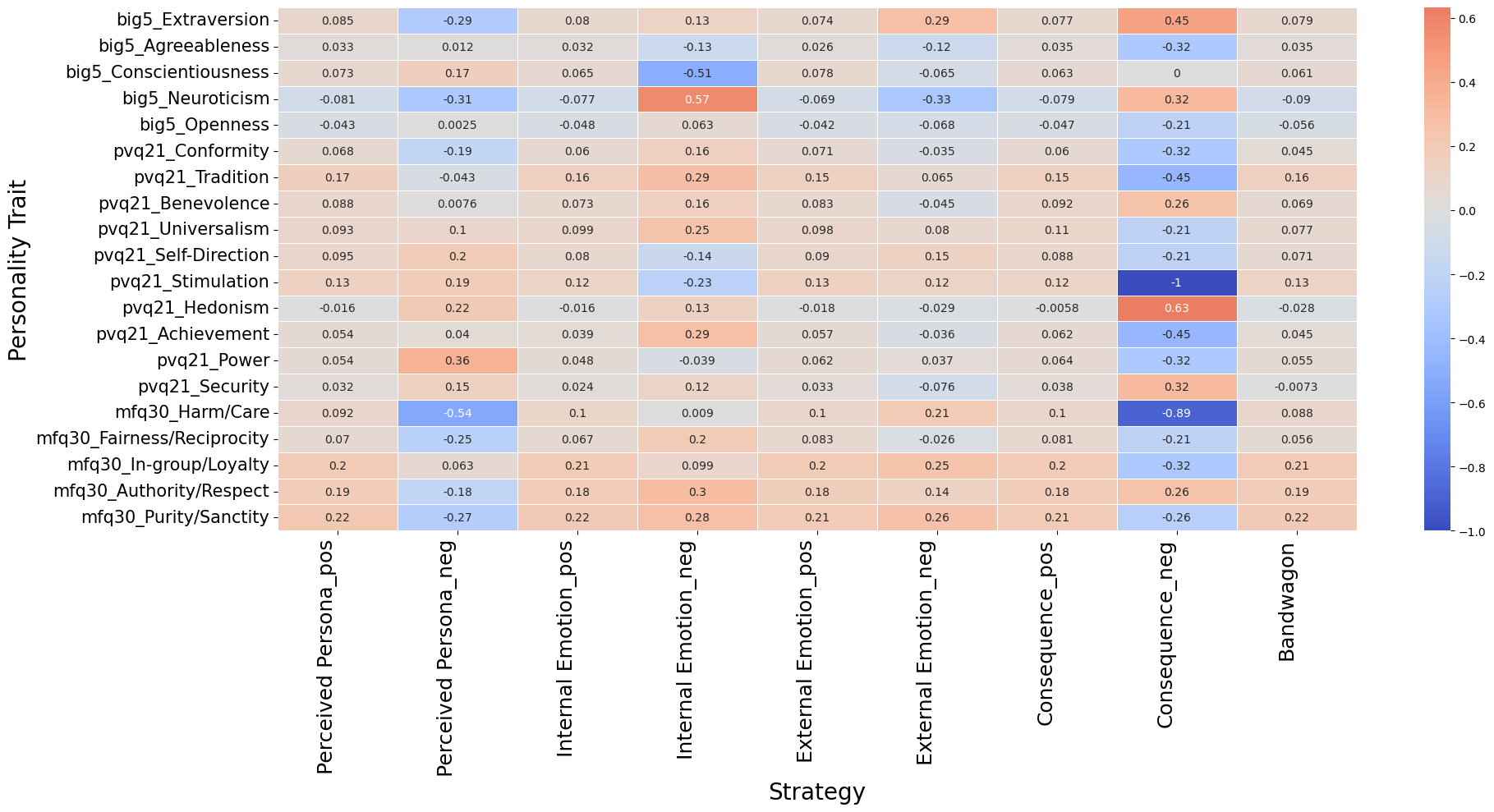}
    \caption{Correlation between image persuasiveness and psychological characteristics across different strategies: These plots illustrate the correlation coefficients between image persuasiveness and various characteristics scores across multiple strategic approaches.}
    \label{fig:correlation_trait}
\end{figure*}

\begin{figure}[hbt!]
    \centering
    \includegraphics[width=1\linewidth]{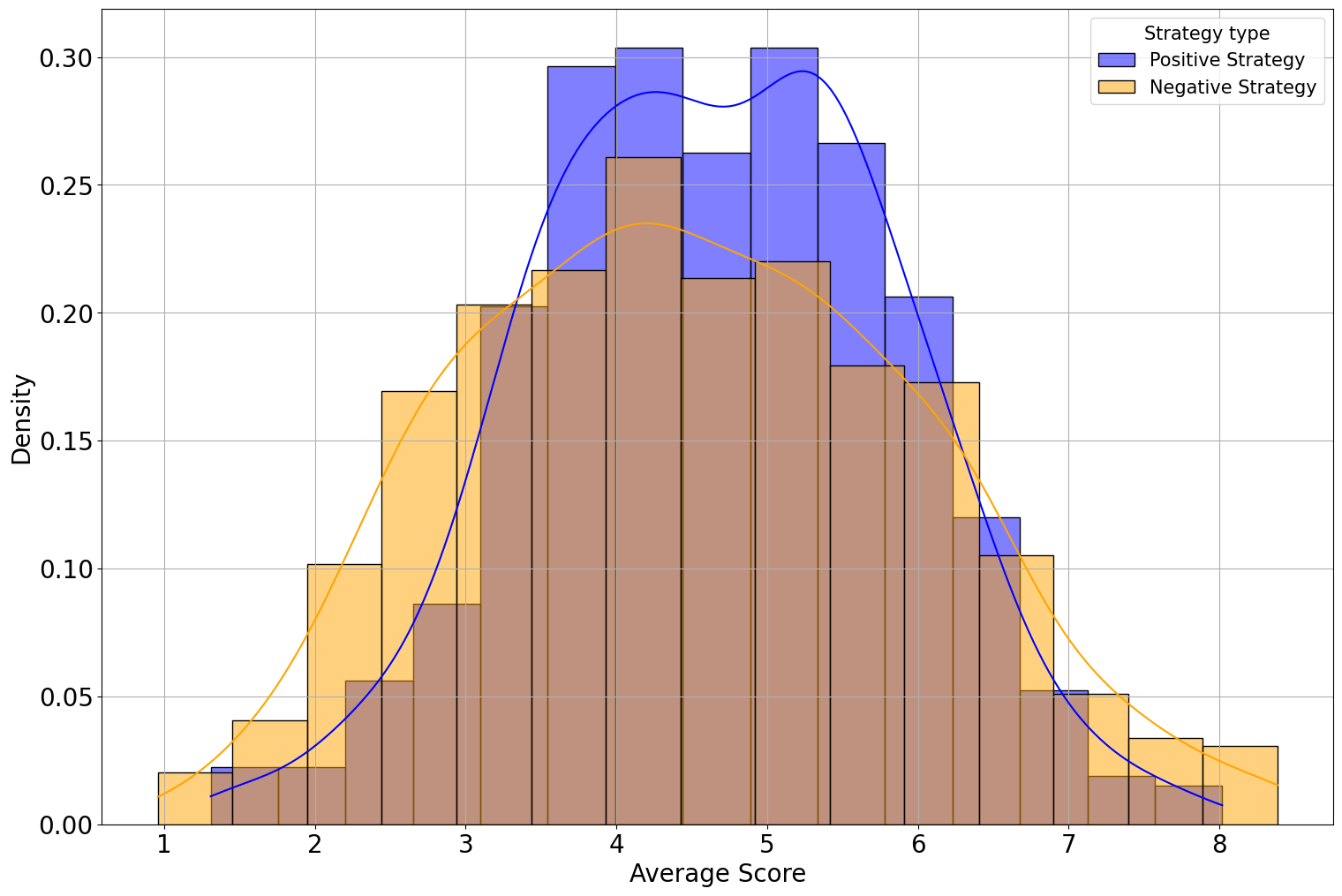}
        \caption{The distribution of average persuasiveness scores for positive and negative strategies. The blue histogram and kernel density estimate represent the average scores for the positive strategy, while the orange histogram and kernel density estimate represent the average scores for the negative strategy.}
    \label{fig:strategy_type}
\end{figure}

\begin{figure}[hbt!]
    \centering
    \includegraphics[width=1\linewidth]{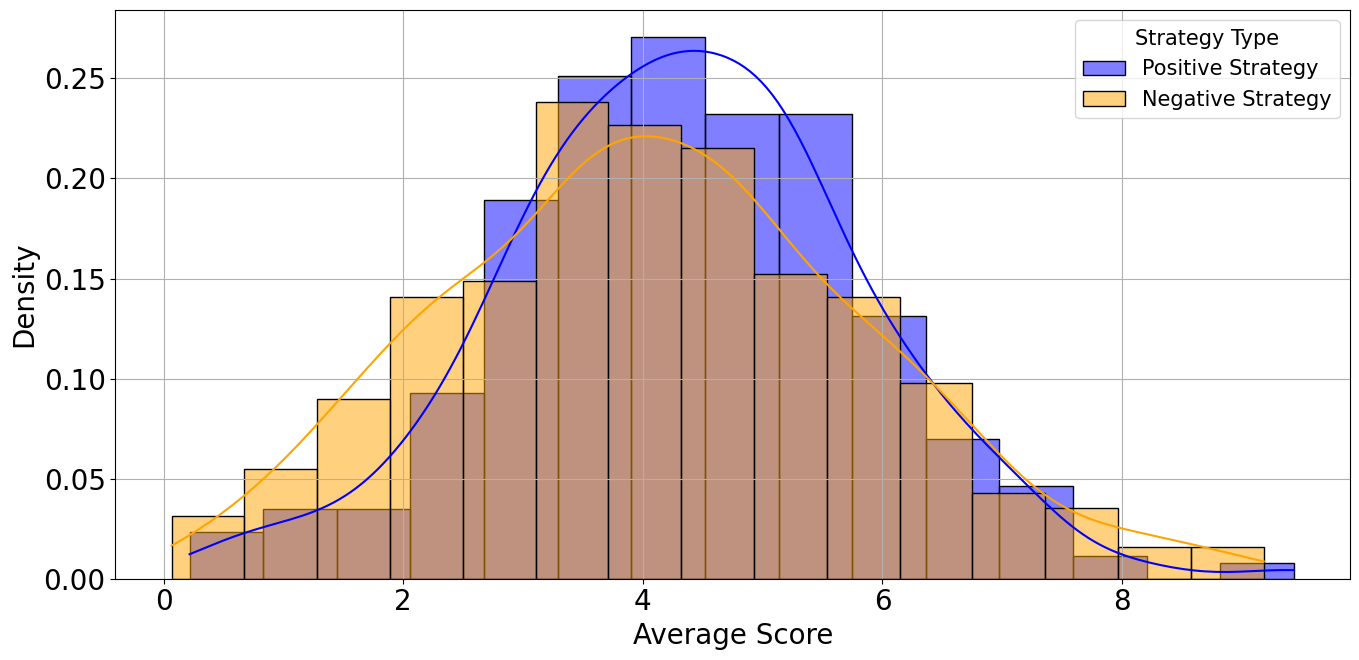}
        \caption{The distribution of average persuasiveness scores for positive and negative strategies on non-habitual individuals. The blue histogram and kernel density estimate represent the average scores for the positive strategy, while the orange histogram and kernel density estimate represent the average scores for the negative strategy.}
    \label{fig:habit}
\end{figure}

\begin{figure}[hbt!]
\centering
\includegraphics[width=1\linewidth]{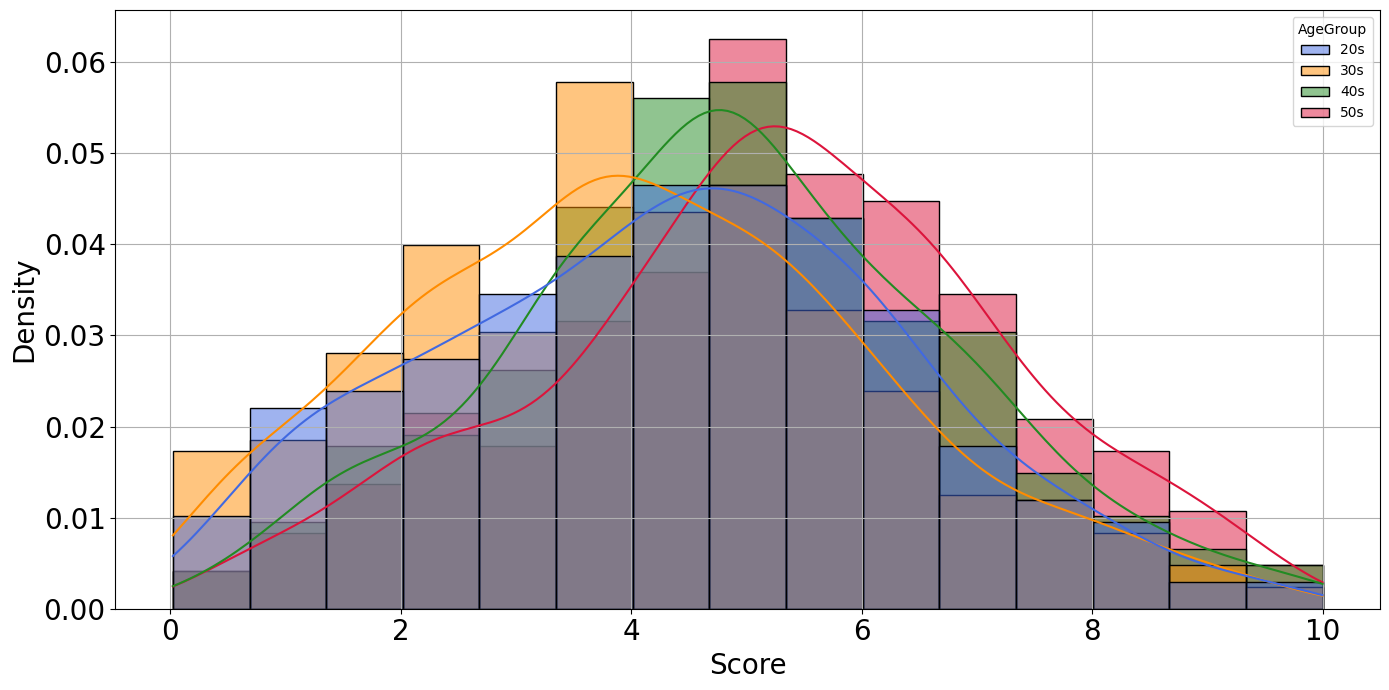}
\caption{The Score Distribution of Ages. The blue color represents people in their 20s, the yellow color represents people in their 30s, the green color represents people in their 40s, and the red color represents people in their 50s.}
\label{fig:Age}
\end{figure}

\begin{figure}[hbt!]
\centering
\includegraphics[width=1\linewidth]{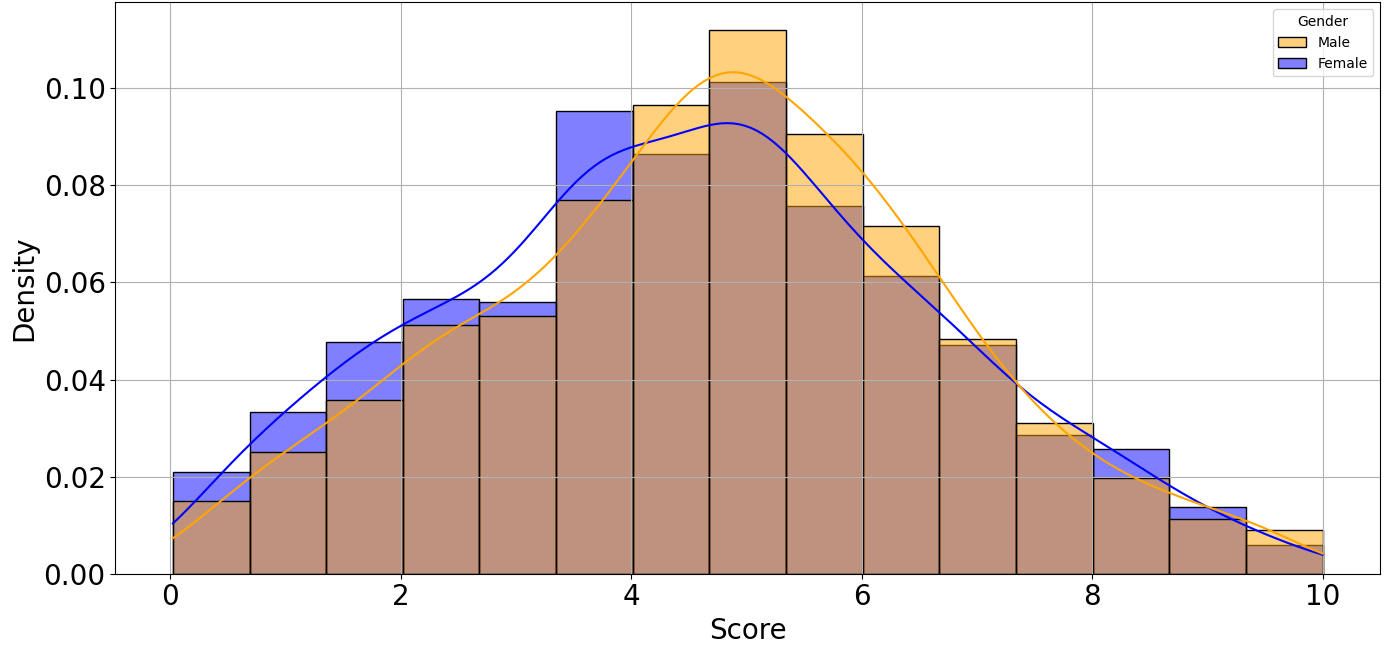}
\caption{Distribution of persuasiveness scores by gender. The yellow histogram and kernel density estimate represent the score distribution for males, while the blue histogram and kernel density estimate represent the score distribution for females.}
\label{fig:Gender}
\end{figure}

\begin{figure}[hbt!]
    \centering
    \includegraphics[width=1\linewidth]{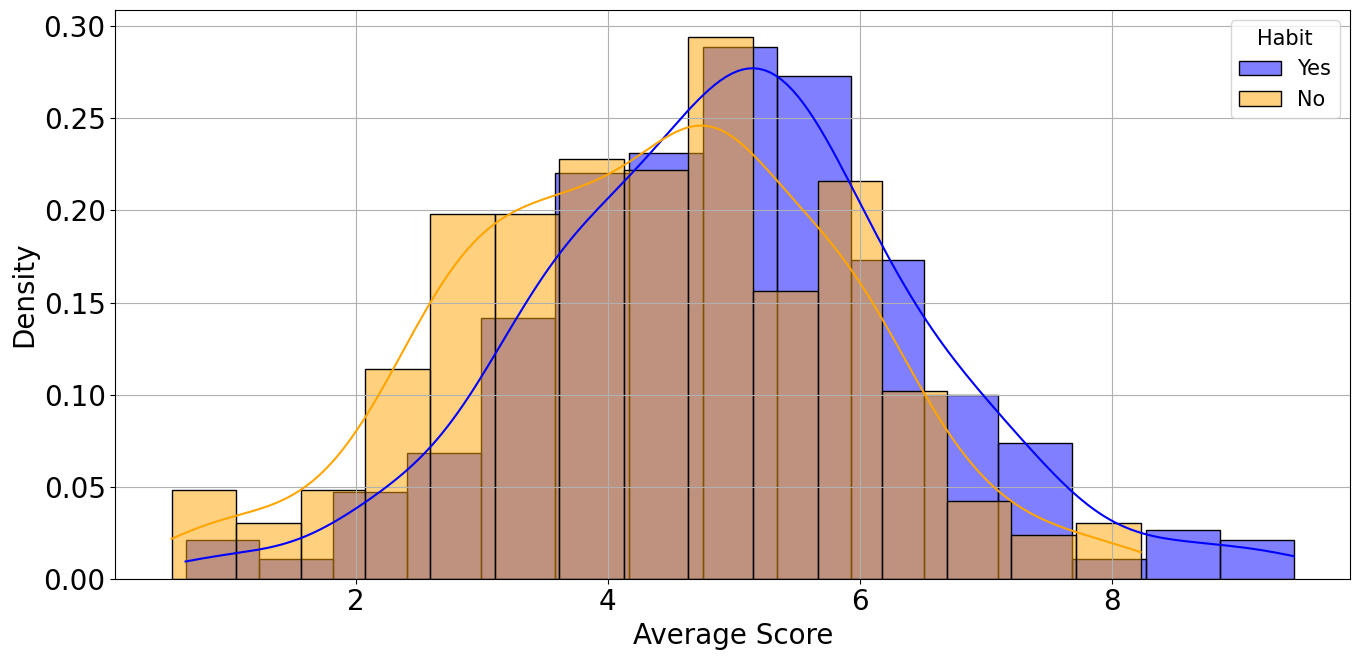}
        \caption{Distributions of average persuasiveness scores by those who had adopted the target behaviors (Habit=Yes) versus those who had not (Habit=No). The blue color represents Yes, and the yellow color represents No.}
    \label{fig:habit_yes_no.png}
\end{figure}

\begin{figure}[hbt!]
    \centering
    \includegraphics[width=1\linewidth]{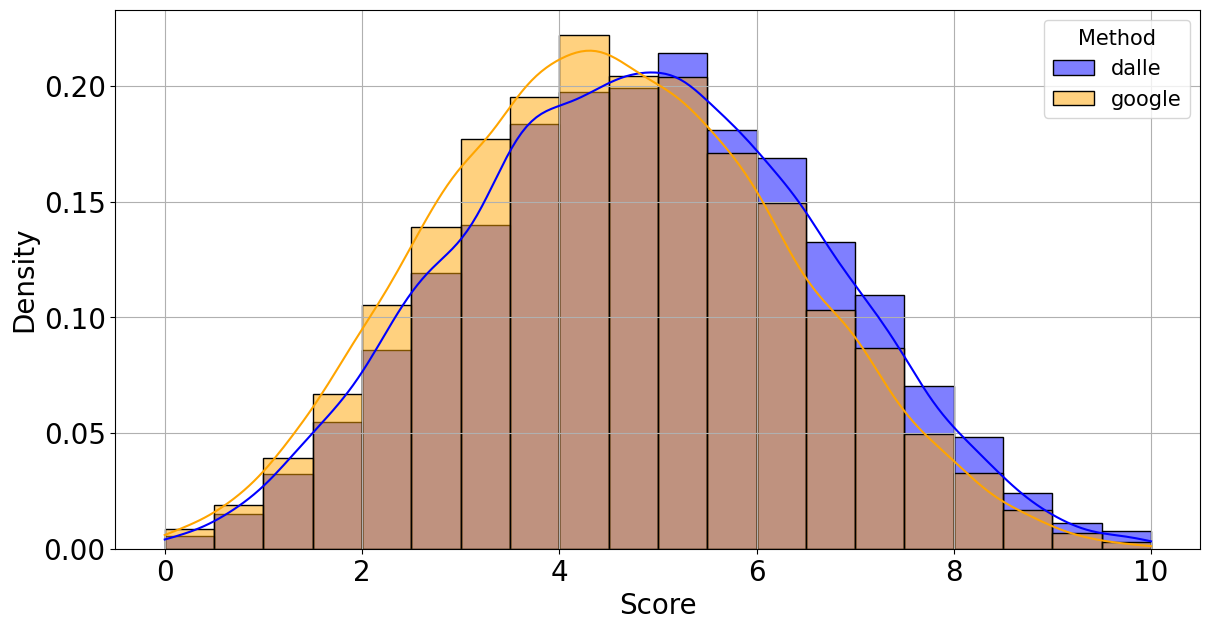}
    \caption{Distributions of persuasiveness scores of images generated by DALLE and images collected through Google Search images. 
    The blue color represents DALLE, and the yellow color represents Google Search images.
    }
    \label{fig: Dalle_Google}
\end{figure}

\section{Evaluator Details} \label{app:Evaluator}


\subsection{Prompts for Evaluator with Four Types of Input Characteristics}
\label{app:Prompts for Evaluator with Four Types of Input Characteristics}

We used prompts that were adapted for four different types of psychological characteristics inputs. Additionally, as the assistant role, ``\#\#\# Response:'' was consistently added after the user prompt during training and inference.

\textbf{PVQ-21 Model Prompt}

\begin{lstlisting}[caption={Prompt 12: System prompt for evaluator using PVQ-21},  label={lst: System prompt for evaluator using PVQ-21}]
You are an AI assistant with expertise in psychology and sociology, specializing in Schwartz's Theory of Basic Values. Your role is to analyze images and messages, evaluating their persuasiveness based on given value priorities.
\end{lstlisting}
\begin{lstlisting}[caption={Input example of PVQ-21},  label={lst: Input example of PVQ-21}]
{"Conformity": 4.0, "Tradition": 4.0, "Benevolence": 4.5, "Universalism": 4.0, "Self-Direction": 4.0, "Stimulation": 4.0, "Hedonism": 4.5, "Achievement": 4.0, "Power": 4.0, "Security": 4.0}
\end{lstlisting}
\begin{lstlisting}[caption={Prompt 13: User prompt for evaluator using PVQ-21}, label={lst: User prompt for evaluator using PVQ-21}]
Below is an instruction that describes a task, paired with an input that provides further context. Write a response that appropriately completes the request.


### Instruction:

You will perform a task where you predict how persuasive certain individuals will find an image created from a message, rating it from 0 to 10. Please predict the persuasiveness score based on the image description and the user's values. These values are based on Schwartz's 10 basic values, where each value is rated on a scale from 1 to 6. The higher the value, the more emphasis is placed on that value. Respond with a single number between 0 and 10.

### Input:

Message: {message}
Value: {pvq21}
Image Description: {image_description}

please directly output a score by strictly following this format: [[score]], for example: [[4]]
\end{lstlisting}

\textbf{MFQ-30 Model Prompt}

\begin{lstlisting}[caption={Prompt 14: System prompt for evaluator using MFQ-30},  label={lst: System prompt for evaluator using MFQ-30}]
You are an AI assistant with expertise in moral psychology, specializing in the Moral Foundations Theory (MFQ). Your role is to analyze images and messages, evaluating their persuasiveness based on moral foundations.
\end{lstlisting}
\begin{lstlisting}[caption={Input example of MFQ-30},  label={lst: Input example of MFQ-30}]
{"Harm/Care": 16, "Fairness/Reciprocity": 14, "In-group/Loyalty": 12, "Authority/Respect": 13, "Purity/Sanctity": 14}
\end{lstlisting}
\begin{lstlisting}[caption={Prompt 15: User prompt for evaluator using MFQ-30}, label={lst: User prompt for evaluator using MFQ-30}]
Below is an instruction that describes a task, paired with an input that provides further context. Write a response that appropriately completes the request.

### Instruction:
You will perform a task where you predict how persuasive certain individuals will find an image created from a message, rating it from 0 to 10. Please predict the persuasiveness score based on the image description and the user's Moral Foundation scores. These scores are based on the Moral Foundations Questionnaire 30 (MFQ30), and each score reflects the individual's importance placed on each moral foundation domain. Respond with a single number between 0 and 10.

### Input:
Message: {message}
MFQ: {mfq30}
Image Description: {image_description}

please directly output a score by strictly following this format: [[score]], for example: [[4]]
\end{lstlisting}

\textbf{Big5 Model Prompt}

\begin{lstlisting}[caption={Prompt 16: System prompt for evaluator using Big5},  label={lst: System prompt for evaluator using Big5}]
You are an AI assistant with expertise in psychology, specializing in the Big Five personality traits. Your role is to analyze images and messages, evaluating their persuasiveness based on the Big Five personality dimensions.
\end{lstlisting}
\begin{lstlisting}[caption={Input example of Big5},  label={lst: Input example of Big5}]
{"Extraversion": 7, "Agreeableness": 7, "Conscientiousness": 5, "Neuroticism": 7, "Openness": 6}
\end{lstlisting}
\begin{lstlisting}[caption={Prompt 17: User prompt for evaluator using Big5}, label={lst: User prompt for evaluator using Big5}]
Below is an instruction that describes a task, paired with an input that provides further context. Write a response that appropriately completes the request.

### Instruction:
You will perform a task where you predict how persuasive certain individuals will find an image created from a message, rating it from 0 to 10. Please predict the persuasiveness score based on the image description and the individual's Big 5 personal traits, where higher scores reflect stronger manifestations of the associated behaviors and emotions, with each trait being scored between 2 and 10. Respond with a single number between 0 and 10.

### Input:
Message: {message}
Big5: {big5}
Image Description: {image_description}

please directly output a score by strictly following this format: [[score]], for example: [[4]]
\end{lstlisting}

\textbf{None Model Prompt}

\begin{lstlisting}[caption={Prompt 18: System prompt for ``None'' evaluator},  label={lst: System prompt for ``None'' evaluator}]
You are an AI assistant with expertise in analyzing and evaluating the persuasiveness of images and messages based on general principles of communication and psychology.
\end{lstlisting}
\begin{lstlisting}[caption={Prompt 19: User prompt for ``None'' evaluator}, label={lst: User prompt for "None" evaluator}]
Below is an instruction that describes a task, paired with an input that provides further context. Write a response that appropriately completes the request.

### Instruction:
You will perform a task where you predict how persuasive certain individuals will find an image created from a message, rating it from 0 to 10. Please predict the persuasiveness score based on the image description. Respond with a single number between 0 and 10.

### Input:
Message: {message}
Image Description: {image_description}

please directly output a score by strictly following this format: [[score]], for example: [[4]]
\end{lstlisting}

\subsection{Implementation Details for Evaluator} \label{app:Implementation Details for Evaluator}

We trained the evaluator model for 1 epoch using the Supervised Fine-Tuning (SFT) approach, applying QLoRA \cite{NEURIPS2023_1feb8787}. We used HuggingFace's SFTTrainer for training.
We set the initial learning rate to 2e-4, with a per-device batch size of 4 and gradient accumulation steps of 8. To further improve training and inference efficiency, we enabled BF16. We set the random seed to 42 for reproducibility. The PEFT configuration is as follows: \{``lora\_alpha'': 16, ``lora\_dropout'': 0.1, ``r'': 64, ``target\_modules'': ``q\_proj'', ``v\_proj''\}.
All experiments were performed on an A100 GPU, with each training session taking approximately 3 hours.

\subsection{Spearman Correlations by Topic}\label{app:Spearman Correlations by Topic}
\begin{figure*}[hbt!]
    \centering
    \includegraphics[width=1\linewidth]{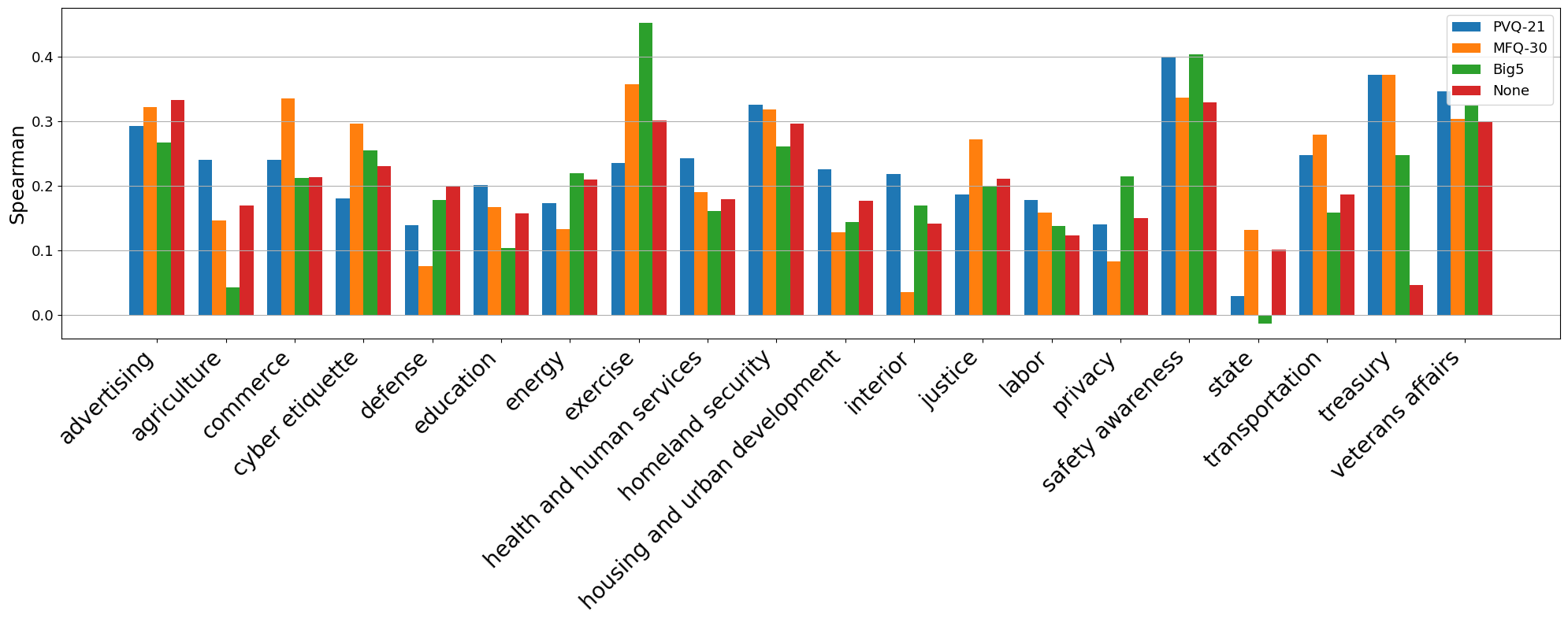}
    \captionof{figure}{Per-topic Spearman correlation between human judgments and four evaluators based on different configurations of psychological characteristics.}
    \label{fig:full_spearman_per_topic}
\end{figure*}

Figure \ref{fig:full_spearman_per_topic}
presents the Spearman correlation between human annotators and four evaluators, each fine-tuned with four different configurations based on different topics. There are a total of 20 topics, and the PVQ-21 model is the most balanced and consistently high-performing model in terms of correlation.
In contrast, the Big5 model and MFQ-30 model exhibit stronger performance in specific domains but lack the overall consistency and stability of the PVQ-21 model.

The None model generally shows lower performance across most topics; however, it achieves relatively high performance in domains like Safety Awareness, which rely on universal and intuitive rules.

\subsection{Spearman Correlations by Strategy}

\begin{figure*}[bbt!]
    \centering
    \includegraphics[width=1\linewidth]{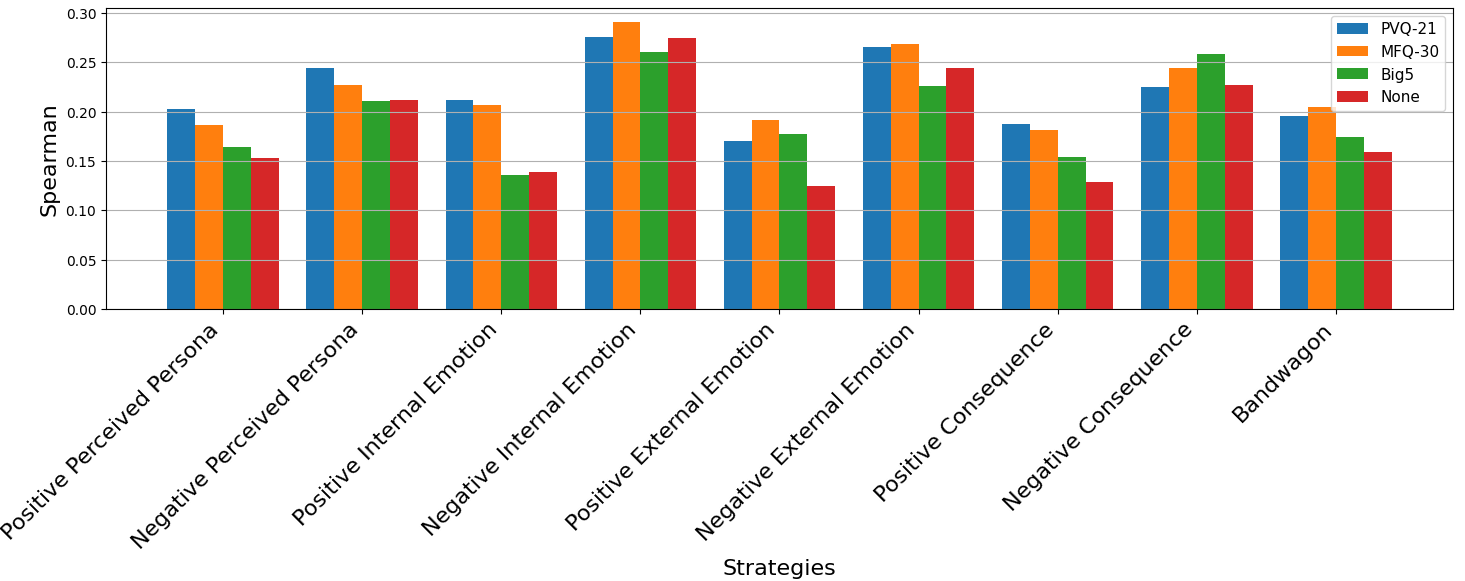}
    \captionof{figure}{Per-strategy Spearman correlation between human judgments and four evaluators based on different configurations of psychological characteristics.}
    \label{fig: metric_per_strategy}
\end{figure*}

Figure \ref{fig: metric_per_strategy}
presents the Spearman correlation between human annotators and four evaluators, each fine-tuned with four different configurations, measured across 9 strategies. The Negative Internal Emotion strategy shows the best performance across all models, while the Positive Consequence strategy and Positive External Emotion strategy generally show worse performance when evaluated by the models.

\subsection{Comparison of Metrics by Google and DALLE} ~\label{app:Comparison of Metrics by Google and DALLE}
Table~\ref{tab:google_dalle_metric} shows evaluator's performance on the Google Image Search and DALLE. Google refers to metrics measured using Google Image Search on the test dataset, while DALLE refers to metrics measured using DALLE on the test dataset. The evaluations in both cases were conducted using the LLaMA3-8B-F evaluator.

\begin{table}[t]
    \small
    \centering
    \begin{tabularx}{\linewidth}{@{}Xcccc}
        \toprule
         & Spearman$\uparrow$ & Pearson$\uparrow$ & NDCG$\uparrow$ & RMSE$\downarrow$ \\ \midrule
         Dalle    & \textbf{0.25}    & \textbf{0.25}    & \textbf{0.53} & 3.44  \\
        Google & 0.24 & 0.24 & 0.49 & \textbf{3.35}\\
         \bottomrule
    \end{tabularx}
    \caption{Evaluator performance on the Google Image Search and DALLE}
    \label{tab:google_dalle_metric}
\end{table}

\subsection{Generalizability of Evaluators to Unseen Messages}
\label{app:generalizability_of_evaluators_to_unseen_messages}
To evaluate the generalizability of the evaluators to unseen messages, we conducted additional experiments where six messages from each topic were randomly selected for testing and the remaining messages were used for training. Table~\ref{tab:unseen_seen} presents the performance of the evaluators in this setting, compared to the best-performing model in the original setting as reported in Section \ref{app:Experiments_main}.

The result suggests that the fine-tuned model generalizes to unseen messages comparably well to the original setting, where images were shared between the training and test sets without annotator overlap. This finding also indicates that predicting persuasiveness scores for unseen annotators, even for previously seen images, is equally challenging.

\begin{table}[h!]
\small
\centering
\begin{tabularx}{\linewidth}{@{}Xcccc}
\toprule
 & Spearman$\uparrow$ & Pearson$\uparrow$ & NDCG$\downarrow$ & RMSE$\uparrow$ \\
\midrule
New      & \textbf{0.30} & \textbf{0.30} & 0.35 & \textbf{3.59} \\
Original  & 0.25 & 0.25 & \textbf{0.42} & 3.40 \\
\bottomrule
\end{tabularx}
\caption{Comparison of evaluator performance between the original setting and a new setting}
\label{tab:unseen_seen}
\end{table}

\subsection{Prompt for Multi-modal Evaluator}
We used the following prompt to evaluate the multi-modal setting of GPT-4o and mini evaluator where the input image itself is used as input.
\begin{lstlisting}[caption={Prompt 20: Prompt for GPT-4o/mini multi-modal evaluator using PVQ-21}, label={lst: GPT-4o/mini prompt for multi-modal evaluator using PVQ-21}]
You are a helpful assistant designed to output JSON.

{img}

Below is an instruction that describes a task, paired with an input that provides further context. Write a response that appropriately completes the request.

### Instruction:
You have been given a message and an image, along with information about your general behavior regarding the message. Assuming you do not normally engage in this activity, evaluate each image and rate your willingness to follow the message on a scale from 0 to 10. Consider the provided images and information to justify your rating based on the given values. These values are based on Schwartz's 10 basic values, where each value is rated on a scale from 1 to 6. The higher the value, the more emphasis is placed on that value. Respond with a single number between 0 and 10.

### Input:
Message: {message}
Value: {value}
Respond with a single number between 0 and 10 in the "score" key.
\end{lstlisting}

\section{Generator Details}
\subsection{Implementation Details for Generator}\label{app:Implementation Details for Generator}
\label{subsubsec: implementation details for generator}
Similar to our approach with the evaluator model, we trained the generator model using Supervised Fine-Tuning (SFT) with QLoRA in 5 epoch. The initial learning rate was set to 2e-4, with a per-device batch size of 4 and gradient accumulation steps of 8. To improve training and inference efficiency, we enabled BF16. The random seed and PEFT configuration are the same as our evaluator's. All experiments were performed on an A100 GPU, with each training session taking approximately 1 hours.

\subsection{Prompts for Generator} \label{app:Prompts for Generator}

The prompts used for training and inference with the generator are as follows. Similar to the prompts used for our evaluator, ``\#\#\# Response:'' was added after the user prompt as the assistant role.

\begin{lstlisting}[caption={Prompt 21: System prompt for generator},  label={lst: System prompt for generator}]
You are an helpful AI assistant for generating image description.
\end{lstlisting}
\begin{lstlisting}[caption={Prompt 22: User prompt for generator}, label={lst: User prompt for generator}]
Below is an instruction that describes a task, paired with an input that provides further context. Write a response that appropriately completes the request.

### Instruction:
Generate an image description based on the following task.  You have received a message and an individual's values as input. These values are based on Schwartz's 10 basic values, rated from 1 to 6, with higher scores indicating greater importance to the individual. Craft an image description that conveys the message's intent using only visual elements like colors, symbols, or scenarios that resonate with the individual's values.  Do not include any references to visible text, such as banners, signs, or posters with wording. The description should rely solely on non-verbal cues and should not exceed 10 sentences.


### Input:
Message: {message}
Value: {pvq21}

Just directly output the image description without adding any prefixes or other modifiers.
\end{lstlisting}

\subsection{Generator Output Examples} ~\label{app:comparison of image descriptions}
Table~\ref{tab:generator example} 
presents example image descriptions generated by three different models: GPT-4o, GPT-4o-mini, and LLaMA3-8b-Instruct-Finetuned.

\subsection{Error Analysis of the Generator}~\label{app:Error Analysis of the Generator}
To analyze the reasons why our model, LLaMA3-8B-F generator, generates image descriptions with low scores, we reviewed the reasons provided by the LLaMA3-8B-F evaluator for image descriptions that received low scores (0, 1, or 2).

Prompt 23 refers to the GPT prompt used for this analysis. The categories are broadly divided into four groups:

\begin{itemize}
    \item \textbf{Category 1. Understanding of Psychological Characteristics:} The generator model failed to adequately understand certain features of the psychological characteristics (Values).
    \item \textbf{Category 2. Accuracy of psychological characteristics in the Image Description:} The psychological characteristics are inaccurately reflected in the image description.
    \item \textbf{Category 3. Clarity and Complexity of the Image Description:} The image description is overly simple, vague, or excessively complex.
    \item \textbf{Category 4. Alignment of Image Description with the Message:} The image description fails to effectively represent the intended message.
    \item \textbf{Category 5. Others:} If the above category does not exist (please provide the reason as well).
\end{itemize}

A total of 116 low-scoring image descriptions were selected for analysis. For each analysis, two major evaluation criteria were chosen. However, in some cases, only a single evaluation criterion was applied to certain image descriptions.

Table~\ref{tab:error analysis} shows categories and corresponding counts of evaluation criteria used for analyzing low-scoring image descriptions. ``Alignment of Image Description with the Message'' was the most frequently selected criterion.

\begin{table}[t]
\resizebox{\columnwidth}{!}{%
\begin{tabular}{ll}
\hline
Category                                           & Counts    \\ \hline
Understanding of Psychological
Characteristics & 86 \\
Accuracy of psychological characteristics in the Image Description & 4 \\
Clarity and Complexity of the
Image Description & 0 \\
Alignment of Image Description
with the Message & 116 \\
Others & 0

\\ \bottomrule

\end{tabular}%
}
\caption{Categories and corresponding counts of evaluation criteria used for analyzing low-scoring image descriptions.}  
\label{tab:error analysis}
\end{table}

\begin{lstlisting}[caption={Prompt 23: Error Analysis of the Generator}]
### Instruction:
You have performed a task where you predicted how persuasive certain individuals would find an image created from a message, rating it from 0 to 10. Each image has been evaluated on a scale from 0 to 10 based on its persuasiveness. The provided data consists of images that received low ratings, ranging from 0 to 2. Based on the image descriptions, the user's values, and the given image descriptions, Please specify which category from the provided Evaluation Criteria the given reason belongs to. These values are based on Schwartz's 10 basic values, where each value is rated on a scale from 1 to 6. The higher the value, the more emphasis is placed on that value.


### Input:
Message: {message}
Value: {value}
Image description: {image_description}
Pesuasiveness Score: {score}
Reason: {reason}


### Evaluation Criteria: 

Please select two major types of Evaluation Criteria:

Category1. Understanding of Psychological Characteristics: The generator model failed to adequately understand certain features of the psychological characteristics (Values).
Category2. Accuracy of psychological characteristics in the Image Description: The psychological characteristics are inaccurately reflected in the image description.
Category3. Clarity and Complexity of the Image Description: The image description is overly simple, vague, or excessively complex.
Category4. Alignment of Image Description with the Message: The image description fails to effectively represent the intended message.
Category5. Others: Please include any additional reasons that do not fit into the specified categories.

Please directly output a reason by strictly following this format: [[Type of Evaluation Criteria: Evaluation Criteria Category Num], [Type of Evaluation Criteria: Evaluation Criteria Category Num]], for example: [[Type of Evaluation Criteria: 1], [Type of Evaluation Criteria: 2]]
\end{lstlisting}

\begin{table*}[hbt!]
\centering
\small
\renewcommand{\arraystretch}{1.5}
\begin{tabularx}{\linewidth}{@{}p{2.5cm}p{1.8cm}X@{}}    
\toprule
    Model & Type & Description \\ 
\midrule 
\multirow{1}{*}{GPT-4o} & Input  & Message: Do\_not\_watch\_TV\_while\_studying  Value: \{``Achievement'': 5.0, ``Benevolence'': 5.5, ``Conformity'': 4.5, ``Hedonism'': 5.0, ``Power'': 4.5, ``Security'': 4.0, ``Self-Direction'': 5.0, ``Stimulation'': 5.0, ``Tradition'': 3.0, ``Universalism'': 4.3\} \\ 
           & Output & In a cozy study room, an open notebook and a pen are centered on a wooden desk, surrounded by warm light from a desk lamp, creating an inviting and peaceful atmosphere. To the side of the desk, a television is turned off, with a slight dusting on the screen, signaling its infrequent use. In the background, a window reveals a serene view of a lush garden with blossoming flowers and a vibrant rainbow, symbolizing creativity, harmony, and spontaneity. A diverse group of books sits neatly on a nearby shelf, their colorful spines facing outwards to suggest self-directed learning and the joyous energy of exploration. Framed photographs on the wall show a supportive community and meaningful achievements, embodying the values of benevolence and accomplishment. Near the desk, a plant thrives on the windowsill, its leaves exhibiting dynamic growth, reflecting stimulation and a zest for life. The entire space exudes a balance between order and freedom, emphasizing security and individual purpose.\\ 
\midrule 
\multirow{1}{*}{GPT-4o-mini} & Input  & Message: Do\_not\_watch\_TV\_while\_studying Value: \{``Achievement'': 5.0, ``Benevolence'': 5.5, ``Conformity'': 4.5, ``Hedonism'': 5.0, ``Power'': 4.5, ``Security'': 4.0, ``Self-Direction'': 5.0, ``Stimulation'': 5.0, ``Tradition'': 3.0, ``Universalism'': 4.3\} \\ 
           & Output & A cozy study room bathed in soft, natural light, with a sturdy wooden desk at its center. On the desk, an open notebook filled with neatly written notes accompanied by colorful pens, symbolizing the pursuit of knowledge and achievement. A serene plant, lush and vibrant, sits in the corner, embodying benevolence and universalism through its connection to nature. A closed laptop rests to the side, subtly suggesting a break from distractions, while a small clock ticks softly, indicating the value of time management. A window reveals a tranquil outdoor scene, featuring trees and birds, reflecting the joy of stimulation and hedonic experiences outside the study space. The atmosphere is peaceful yet invigorating, encouraging a sense of self-direction and motivation. Throughout the room, warm colors dominate, creating a welcoming environment that fosters focus and curiosity. A motivational trophy, shining on a shelf, stands as a testament to past achievements and future aspirations. The overall composition encourages a balance between personal growth and the fulfillment of responsibilities, free from the temptation of television.\\ 
\midrule 
\multirow{2}{*}{\shortstack{LLaMA3-8b-\\Instruct-Finetuned}} & Input  & Message: Do\_not\_watch\_TV\_while\_studying Value: \{``Achievement'': 5.0, ``Benevolence'': 5.5, ``Conformity'': 4.5, ``Hedonism'': 5.0, ``Power'': 4.5, ``Security'': 4.0, ``Self-Direction'': 5.0, ``Stimulation'': 5.0, ``Tradition'': 3.0, ``Universalism'': 4.3\} \\ 
           & Output & The image depicts a young woman sitting at a cluttered desk in a dimly lit room. She appears to be stressed, with her hands on her head and a worried expression on her face. The desk is covered with various items including books, a laptop, and scattered papers. A TV screen behind her shows a news broadcast with a serious headline reading 'STUDYING VS. TV: THE NEGATIVE EFFECTS' and a picture of a person studying. The room is dimly lit, with a single desk lamp casting a warm glow on the scene, and a window with blinds partially open in the background, allowing some natural light to filter in. The overall atmosphere suggests a sense of tension and concern about the impact of television on studying. The woman is dressed in a casual outfit, wearing a dark top and blue jeans. There is a clock on the wall showing the time as 9:00, adding to the sense of urgency. The floor is littered with more papers and books, contributing to the chaotic environment. The image conveys a message about the importance of prioritizing studying over TV.\\ 
\bottomrule
\end{tabularx}
\caption{Examples of input and output descriptions from different generators.}
\label{tab:generator example}
\end{table*}

\section{Survey of Persuasiveness Score}
To obtain persuasiveness scores for each image, we designed survey forms tailored to each message. The survey consists of three types of questionnaires: 1) Persuasiveness Score, 2) Habit, and 3) Psychological Characteristics.

The following is an example of 1), and 2) from the survey for the message, ``Fasten your seatbelt.''

\begin{lstlisting}[caption={Survey Instructions},  label={lst: Instruction of survey}]
# Instruction:

You have been given the message, `Fasten your seatbelt' Assuming that you usually do not fasten your seatbelt, please look at each image and rate how much the image makes you want to fasten your seatbelt on a scale from 0 to 10. (0: Not at all, 10: Very strongly)

# Persuasiveness Score:

Message: Fasten your seatbelt.

{img}

Score:
- 0 [ ] 
- 1 [ ] 
... 
- 9 [ ] 
- 10 [ ]

# Habit:

Do you always fasten your seatbelt?
- Yes [ ]
- No [ ]



\end{lstlisting}

\end{document}